\pdfoutput=1

\documentclass{article}

\usepackage{microtype}
\usepackage{graphicx}
\usepackage{subcaption}
\usepackage{booktabs} %

\usepackage{hyperref}

\usepackage[accepted]{icml2026}

\usepackage{amsmath}
\usepackage{amssymb}
\usepackage{mathtools}
\usepackage{amsthm}

\usepackage{algorithm}
\usepackage{algorithmic}
\usepackage{url}
\usepackage{latexsym}

\usepackage{xcolor}

\usepackage[capitalize,noabbrev]{cleveref}

\usepackage{tikz}
\usetikzlibrary{automata, positioning, arrows.meta}

\usepackage{enumitem}
\setitemize{noitemsep,topsep=0pt,parsep=0pt,partopsep=0pt}

\theoremstyle{plain}

\theoremstyle{definition}

\theoremstyle{remark}

\usepackage{CJKutf8}
\usepackage{inconsolata}

\icmltitlerunning{Beyond Perplexity: UTF-8 Validity in Byte-aware Language Models}

\begin{document}

\twocolumn[
\icmltitle{Beyond Perplexity: UTF-8 Validity in Byte-aware Language Models}

\icmlsetsymbol{equal}{*}

\begin{icmlauthorlist}
\icmlauthor{Sangwhan Moon}{google}
\icmlauthor{Daisuke Oba}{isct}
\icmlauthor{Youmi Ma}{isct}
\icmlauthor{Tatsuya Hiraoka}{mbzuai}
\icmlauthor{Naoaki Okazaki}{isct}
\end{icmlauthorlist}

\icmlaffiliation{google}{Google LLC / Mountain View, CA, USA}
\icmlaffiliation{isct}{Institute of Science Tokyo / Tokyo, Japan}
\icmlaffiliation{mbzuai}{Mohamed bin Zayed University of Artificial Intelligence / Abu Dhabi, United Arab Emirates}

\icmlcorrespondingauthor{Sangwhan Moon}{sangwhan@iki.fi}

\icmlkeywords{Machine Learning, ICML, Byte Sequence Modeling, Scaling Laws}

\vskip 0.3in
]

\printAffiliationsAndNotice{}  %

\begin{abstract}
Byte-level tokenization enables language models to handle any Unicode input,
but models can generate invalid UTF-8 sequences when encountering rare or
unseen characters. We investigate the relationship between training scale
and UTF-8 generation reliability with a 355M parameter model trained on 80B
tokens from a balanced multilingual corpus of English, Japanese, Korean, and
Chinese. We introduce multiple evaluation protocols that isolate UTF-8 structural
validity from language modeling. UTF-8 validity convergence lags perplexity by 
roughly a factor of two: perplexity stabilizes after 2.1B tokens, but UTF-8 validity 
requires 4.2B tokens. In context-free generation, common characters achieve higher
structural validity than rare characters, with byte-length exposure emerging
as an additional axis of difficulty alongside frequency. Our experiments show that
reliable UTF-8 generation is a distinct capability requiring evaluation
beyond perplexity.%
\begin{center}
\vspace{-\topsep}
\href{https://github.com/cynthia/bytecanary}{github.com/cynthia/bytecanary}
\end{center}

\end{abstract}

\section{Introduction}

Multilingual NLP systems inevitably face unknown characters: limited
vocabulary budgets prevent full Unicode coverage, causing tokenizers to fail on
characters outside their vocabulary. This problem often occurs when handling
languages that use non-Latin alphabets, such as CJK languages. To mitigate this, most popular large language models (LLMs) utilize byte-fallbacks that
includes tokens corresponding to all bytes so that the tokenizer can encode texts
with unknown characters. %

Beyond the byte-fallback, some researchers have attempted to develop models with
byte-level tokenization, which allows tokens to be separated by byte boundaries
rather than character boundaries~\cite{gillick-etal-2016-multilingual}. Notable
examples include ByT5~\citep{xue-etal-2022-byt5}, which demonstrated competitive
performance operating directly on UTF-8 bytes, and the Byte Latent Transformer
(BLT)~\citep{custom-pagnoni2024byte}, which achieved comparable results to Llama 3 while using fewer inference FLOPs. The advantage of
byte-level tokenization is that we can flexibly tokenize multi-byte characters
into meaningful minimal units. 
For example, we can extract a primary radical from a single hanzi/kanji character or a single jamo character into pronunciation units. 
This flexibility enables accurate multi-byte character understanding, effective token embedding learning, and improved downstream task performance~\cite{xue-etal-2022-byt5}.

Despite these advantages, byte-level tokenization also has a problem in its
decoding phase. Since the vocabulary in byte-level tokenization allows tokens
that begin or end with a byte in the middle of characters, an NLP system
sometimes generates an invalid Unicode sequence, where some tokens cannot
connect with each other appropriately. This damages performance in NLP tasks
that require models to generate texts, such as machine
translation~\cite{wang2020neural}.
Prior work in the LLM field has not thoroughly discussed this problem, likely
because byte-level tokenization has been deployed primarily in sufficiently
large systems capable of learning valid Unicode sequences. However, considering
the difficulty of learning valid Unicode sequences, we assume that byte-level
tokenization requires significantly more training data or trainable parameters
so that the model can sufficiently learn to generate valid sequences.

We investigate this by training a 355M parameter GPT-2 architecture
on 80B tokens of multilingual data comprising English (10\%) and Japanese,
Korean, and Chinese (30\% each), evaluating UTF-8 validity across 420 training
checkpoints. UTF-8 validity convergence lags perplexity convergence by a factor
of two: perplexity stabilizes after 2.1B tokens, but UTF-8 validity requires
4.2B tokens. Common characters achieve higher structural validity than rare
ones (96.21\% vs.\ 95.26\%); a control experiment further shows that byte-length
exposure is a major axis of difficulty alongside character frequency.
Structural validity also exceeds semantic correctness---Term Match Rate reaches
60.30\% despite high validity---suggesting that generating a \textit{valid}
character is easier than generating the \textit{correct} one.
These results have practical implications: models that appear well-trained by
perplexity may still produce invalid UTF-8 sequences in context-sparse
generation.

\section{Problem Statement}

\begin{figure}[t]
    \centering
    \includegraphics[width=0.95\linewidth]{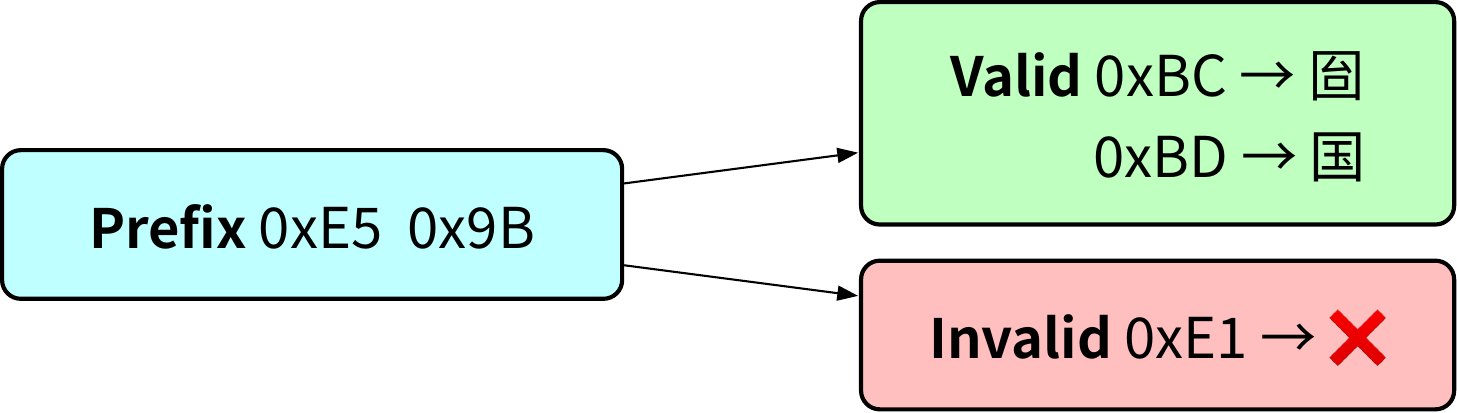}
    \caption{An example of valid vs. invalid byte sequences. The invalid case cannot be decoded by a UTF-8 codec.}
    \label{fig:example_valid_invalid}
\end{figure}

Modern large language models face a fundamental challenge in generating valid
Unicode byte sequences, particularly when encountering rare or unseen
characters. While byte-level tokenization offers theoretical advantages over
character-based approaches---including the ability to flexibly tokenize
multi-byte characters and handle any possible input---it introduces a critical
failure mode: models can generate invalid UTF-8 sequences that violate Unicode
encoding constraints (Figure~\ref{fig:example_valid_invalid}).

To understand this problem, consider how UTF-8 encoding works. Each Unicode
character is encoded as a sequence of one to four bytes following strict
patterns. ASCII characters (U+0000 to U+007F) use a single byte with the pattern
\texttt{0xxxxxxx}. Characters from U+0080 to U+07FF require two bytes following
\texttt{110xxxxx 10xxxxxx}, while characters from U+0800 to U+FFFF need three
bytes as \texttt{1110xxxx 10xxxxxx 10xxxxxx}. Finally, characters from U+10000
to U+10FFFF are encoded with four bytes following \texttt{11110xxx 10xxxxxx
10xxxxxx 10xxxxxx}. Each continuation byte must begin with \texttt{10}, and the
leading byte determines how many continuation bytes follow. When a model trained
with byte-level tokenization generates text, it must implicitly learn these
encoding rules to produce valid sequences. However, this learning depends
critically on exposure to diverse byte patterns during training.

The problem becomes particularly acute in the context of the long-tail character
distribution prevalent in natural language. Consider a rare character like
U+2B740 (CJK ideograph) encoded as \texttt{0xF0 0xAB 0x9D 0x80}. If this
character appears with probability $p(c) < \frac{1}{N}$ (where $N$ is the total
number of tokens in training data), the model may never observe this specific
byte sequence. When prompted to generate text following a prefix containing
\texttt{0xF0 0xAB}, the model must correctly predict that the next byte must
match the pattern \texttt{10xxxxxx} as a continuation byte, i.e.,
\texttt{0x9D} would be valid while \texttt{0xF0} would not, and that after
\texttt{0x9D}, another continuation byte is required. Without sufficient
exposure to similar patterns, the model might generate \texttt{0xF0 0xAB
0xF0}---starting a new 4-byte sequence instead of continuing the current one.
This creates an invalid UTF-8 sequence that cannot be decoded, causing
downstream applications to fail with decoding errors or produce replacement
characters.

The severity of this problem extends beyond simple decoding failures. When
models generate invalid sequences, they can enter unstable states where
subsequent generation becomes increasingly incoherent. We hypothesize that these
failure modes can be triggered adversarially by crafting inputs with rare byte
sequences, potentially causing models to generate streams of invalid bytes that
appear as corrupted text, fall into repetitive patterns trying to ``escape''
invalid states, produce outputs that bypass safety filters due to tokenization
confusion, or exhibit degraded performance on downstream tasks when rare
characters appear.

The core research questions we address are:
\begin{itemize}
\item What is the empirical relationship between training scale and UTF-8 generation reliability?
\item How does UTF-8 validity convergence relate to perplexity convergence during training?
\item Do rare and common characters exhibit different validity learning patterns?
\item Does semantic context affect UTF-8 validity learning?
\end{itemize} 
\section{Evaluation Framework}
\label{sec:eval-framework}

Byte-fallback tokenization allows language models to represent
arbitrary Unicode strings, but generation can still fail by emitting byte
sequences that are not valid UTF-8. Crucially, such failures are only weakly
coupled to standard language modeling metrics: a model can assign high
probability mass to structurally invalid continuations, and perplexity can
stabilize while UTF-8 validity continues to improve. 
We thus propose to evaluate
\emph{UTF-8 generation reliability} as a distinct capability along three dimensions.

Given a prompt $C$ and a generated token sequence $x_{1:T}$, let
$\text{Bytes}(x_{1:T})$ denote the corresponding byte stream after detokenizing
(including bytes produced by byte-fallback tokens). 
We measure:

\textbf{(G1) Structural validity.}
Whether $\text{Bytes}(x_{1:T})$ forms a valid UTF-8 string (and at which step it
fails if not).

\textbf{(G2) Semantic correctness (when a target is defined).}
Whether the model outputs the \emph{correct} byte sequence for a particular OOV character, not merely \emph{some} valid UTF-8.

\textbf{(G3) Probabilistic preference.}
When greedy sampling fails to output the correct bytes, determine whether the model assigns higher likelihood to the correct completion.

We evaluate these axes with two complementary settings, Level~0 and Level~1,
and a metric suite covering structural, semantic, and probabilistic behavior.

\subsection{Evaluation Tasks}
\label{sec:eval-tasks}

\textbf{Level 0: Context-Free Structural Validity.}
Level~0 isolates UTF-8 structural reliability from language understanding.
We prompt the model with prefixes that contain rare or unseen characters
(under byte-fallback), generate continuations, and score the produced byte
streams by UTF-8 validity.
To probe generalization, we stratify targets by character frequency tiers (e.g., common or uncommon).

\textbf{Level 1: Context-Guided Byte Retrieval.}
Level~1 tests whether semantic and syntactic context can guide the model to
retrieve and complete an OOV character's byte sequence. Given a sentence $S$
containing a target character $c$, we form a prefix that includes the context
preceding $c$ and a partial byte prefix of $c$. The model must generate the
remaining bytes of $c$ as the immediate continuation.
This task explicitly couples semantics to byte-level output while keeping the
evaluation local to a short completion window. 

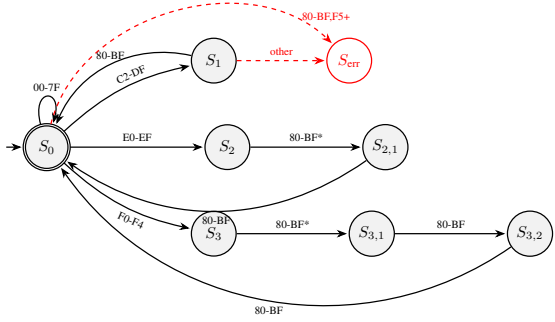
\begin{figure}[t]
\centering
\resizebox{0.9\columnwidth}{!}{%
\begin{tikzpicture}[
    ->,
    >=Stealth,
    shorten >=1pt,
    auto,
    node distance=2.5cm,
    thick,
    scale=0.8,
    main state/.style={circle, draw, thick, fill=gray!10, minimum size=10mm},
    err state/.style={circle, draw=red, thick, minimum size=10mm, text=red},
    lbl/.style={font=\scriptsize, align=center}, 
    initial text={}
]

    \node[state, initial, accepting, main state] (S0) {$S_0$};
    \node[main state] (S1) [above right = 1.2cm and 3cm of S0] {$S_1$};
    \node[main state] (S2) [right = 3cm of S0] {$S_2$};
    \node[main state] (S21) [right = of S2] {$S_{2,1}$};
    \node[main state] (S3) [below right = 1.2cm and 3cm of S0] {$S_3$};
    \node[main state] (S31) [right = of S3] {$S_{3,1}$};
    \node[main state] (S32) [right = of S31] {$S_{3,2}$};
    \node[err state] (Serr) [right = 2cm of S1] {$S_{\text{err}}$};

    \path (S0) edge[loop above] node[lbl] {00-7F} (S0);

    \path (S0) edge[bend left=15] node[lbl, pos=0.6, sloped, above] {C2-DF} (S1);
    \path (S1) edge[bend right=40] node[lbl, above] {80-BF} (S0);

    \path (S0) edge node[lbl, pos=0.5, above] {E0-EF} (S2);
    \path (S2) edge node[lbl, above] {80-BF*} (S21);
    \path (S21) edge[bend left=35] node[lbl, below] {80-BF} (S0);

    \path (S0) edge[bend right=15] node[lbl, pos=0.6, sloped, below] {F0-F4} (S3);
    \path (S3) edge node[lbl, above] {80-BF*} (S31);
    \path (S31) edge node[lbl, above] {80-BF} (S32);
    \path (S32) edge[bend left=45] node[lbl, below] {80-BF} (S0);

    \path (S1) edge[dashed, red] node[lbl, above] {other} (Serr);
    \path (S0) edge[dashed, red, bend left=60] node[lbl, pos=0.85, right] {\ \ 80-BF,F5+} (Serr.north west);

\end{tikzpicture}
}%
\caption{Simplified UTF-8 DFA with dashed red error transitions.} %
\label{fig:utf8-dfa}
\end{figure}

\subsection{Metrics}
\label{sec:metrics}

Our evaluation targets three distinct objectives: 
(G1) \emph{structural} UTF-8 validity of the generated byte stream, 
(G2) \emph{semantic} correctness when a gold target character is defined (Level~1), 
and (G3) \emph{diagnosis} of whether errors arise from missing knowledge or from decoding/calibration. 
No single metric captures all three of them, so we report a complementary suite (Table~\ref{tab:metric-summary}).

\begin{table}[t]
\centering
\small
\caption{Metric roles in our framework.
$\dagger$ Indirect: perplexity reflects average predictive fit, not pairwise preference between gold and generated completions.}
\begin{tabular}{@{}l@{\,\,\,}c@{\,\,\,}c@{\,\,\,}c@{}}
\toprule
& Structural & Semantic  & Diagnostic  \\
Metric & G1 (\S~\ref{sec:utf8-dfa-metric}) & G2 (\S~\ref{sec:term-match}) & G3 (\S~\ref{sec:ll-compare}) \\
\midrule
$V_{\text{partial}}$ (Eq.~\ref{eq:v:partial}) & \checkmark &  &  \\
Term Match (Eq.~\ref{eq:term-match}) &  & \checkmark &  \\
$\Delta_{LL}$ (Eq.~\ref{eq:delta-ll}) &  &  & \checkmark \\
\midrule
Perplexity / NLL &  &  & \checkmark$^{\dagger}$ \\
\bottomrule
\end{tabular}

\label{tab:metric-summary}
\end{table}

\subsubsection{Structural Validity via UTF-8 DFA}
\label{sec:utf8-dfa-metric}

UTF-8 validity is a property of the \emph{detokenized byte stream}.
Let $x_{1:T}$ be the generated token sequence and $B=\text{Bytes}(x_{1:T})$ the corresponding bytes (including byte-fallback tokens). 
Since UTF-8 defines a regular language over bytes, structural validity can be checked exactly by a deterministic finite automaton (DFA; Figure~\ref{fig:utf8-dfa}): 
state $S_0$ represents a character boundary, intermediate states represent partially emitted multi-byte characters awaiting continuation bytes, 
and $S_{\text{err}}$ denotes an invalid transition. 
We enforce standard UTF-8 constraints, rejecting overlong encodings, surrogate halves, and codepoints above U+10FFFF;
see Appendix~\ref{app:utf8-dfa} for full transition details.

\textbf{Partial credit validity.}
Binary validity is brittle when generation stops mid-character (structurally
consistent but incomplete). To separate \emph{incomplete} from \emph{invalid}
outputs, we compute a DFA-based partial-credit score. Let $b_c$ be the number of
bytes belonging to complete valid characters; let $b_i$ be the number of bytes
in the trailing (possibly incomplete) character; and let $p\in[0,1]$ denote the
fractional progress within that trailing character according to the DFA. We define:
\begin{equation}
V_{\text{partial}}(B)=\frac{b_c + p \cdot b_i}{|B|}.
\label{eq:v:partial}
\end{equation}

\textbf{Aggregation across generation steps.}
Per-step structural scores fluctuate during multi-byte emission (e.g., $0.33 \rightarrow 0.67 \rightarrow 1.0$ under $V_{\text{partial}}$ for a 3-byte character). 
For stable monitoring across checkpoints and lengths, we aggregate prefix-wise scores computed on $B_t=\text{Bytes}(x_{1:t})$. 
We report the running mean:
\begin{equation}
V_{\text{cumulative}}(t)=\frac{1}{t}\sum_{i=1}^t V_{\text{partial}}(B_i),
\end{equation}
and, when emphasizing recent behavior, an exponential moving average:
\begin{equation}
V_{\text{ema}}(t)=\alpha V_{\text{partial}}(B_t) + (1-\alpha)V_{\text{ema}}(t-1).
\end{equation}

\subsubsection{Semantic Correctness via Term Match}
\label{sec:term-match}

UTF-8 validity (\S~\ref{sec:utf8-dfa-metric}) does not imply the model produced the \emph{intended} character.
In Level~1, we define a gold byte completion for a target character.
Let the target bytes be $B_c=(B_p,B_r)$, where $B_p$ is the provided byte prefix
and $B_r$ is the remaining suffix to be generated. We report a binary Term Match
indicator that is satisfied only if the model emits exactly $B_r$ as the
immediate continuation \emph{and} returns to a UTF-8 boundary after consuming it:
\begin{equation}
\begin{aligned}
M=\mathbb{I}\big[&
\text{the next bytes complete } B_r \\
&\text{and the DFA state returns to } S_0
\big].
\end{aligned} \label{eq:term-match}
\end{equation}
This excludes cases that are UTF-8 valid but correspond to a different character.

\subsubsection{Diagnosis via Likelihood Comparison}
\label{sec:ll-compare}

To distinguish missing knowledge from decoding/calibration failures, we compare
teacher-forced log-likelihoods of the gold completion versus the generated completion under the same context $C$. For gold token sequence
$X_{\text{gold}}$ and generated sequence $X_{\text{gen}}$, we compute:
\begin{align}
    \Delta_{LL} &= \sum_{t=1}^{|X_{gold}|} \log P_\theta(x_t^{gold} \mid C,
        x_{<t}^{gold}) \nonumber \\
    &\quad - \sum_{t=1}^{|X_{gen}|} \log P_\theta(x_t^{gen} \mid C, x_{<t}^{gen})
    \label{eq:delta-ll}
\end{align}
$\Delta_{LL}>0$ indicates the model prefers the correct completion even when the
decoder fails to emit it (suggesting decoding/calibration issues); $\Delta_{LL}<0$
indicates the model confidently prefers an incorrect continuation.

\subsection{Evaluation Protocols}
\label{sec:eval-protocols}

We evaluate each saved checkpoint on fixed \emph{trial sets} of $M=256$ samples per language for both Level~0 and Level~1, enabling learning curves with minimal evaluation variance. 
Unless otherwise noted, we use the same decoding configuration across checkpoints and compute structural, semantic, and preference metrics as defined in Sec.~\ref{sec:metrics}. 
Full construction details are provided in Appendix~\ref{app:eval-dataset} and \ref{app:gemini-contexts}.

\paragraph{Level 0: frequency-tiered OOV characters.}
We construct a trial set $D_{\text{trial}}$ of OOV characters stratified into four frequency tiers (\emph{Common/Uncommon/Rare/Unseen}) to control difficulty.
We define the set of \emph{seen} characters as $K = V \cup S$, where $V$ is the set of Unicode characters covered by tokenizer vocabulary tokens and $S$ is the set of OOV characters observed in the training corpus under byte-fallback.
The \emph{Unseen} tier is sampled from $U \setminus K$ for a predefined Unicode universe $U$ (details in Appendix~\ref{app:eval-dataset}). 
To enable direct comparability between context-free and context-guided settings, the \emph{Common} tier is chosen to overlap with the Level~1 target pool. 
We use script-aware stratification to avoid mono-script tiers (Appendix~\ref{app:eval-dataset}).

\paragraph{Level 1: context-guided byte completion.}
We extract OOV target characters by scanning the pre-tokenized training stream for contiguous byte-fallback sequences and decoding them to UTF-8.
Level~1 evaluation focuses on \emph{Common} tier characters; synthetic context generation for \emph{Rare} and \emph{Unseen} characters proved unreliable because target usage in natural-sounding sentences was difficult to validate automatically.
To construct controlled contexts without reusing pre-training text, we generate single-sentence prompts using Gemini~3~Pro and filter them for language correctness and uniqueness (Appendix~\ref{app:gemini-contexts}).
Given a sentence containing target character $c$, we apply the \emph{Sentence Prompt} constraint by providing the preceding context $C_{\text{ctx}}$ and a short byte prefix $B_p$ of the target bytes $B_c$; 
the model is evaluated on whether it emits the remaining suffix $B_r$ as the immediate continuation (Sec.~\ref{sec:term-match}).

\section{Experimental Setup}
\label{sec:setup}

\subsection{Model and Tokenizer}
We train a 355M-parameter decoder-only Transformer based on a GPT-2-style
architecture. Implementation details of the model are explained in Appendix \ref{app:arch-budget}.
We use an 8{,}000-token BPE vocabulary with byte-fallback for out-of-vocabulary
characters. When a character is not covered by the vocabulary, it is encoded as
its UTF-8 byte sequence using dedicated byte tokens (e.g.,
\texttt{<0xE4><0xB8><0xAD>} for
\begin{CJK}{UTF8}{min}中\end{CJK}). This preserves information for arbitrary
Unicode input at the cost of sequence length.

\subsection{Training Data}
We construct a multilingual corpus from FineWeb~\cite{fineweb} for English and FineWeb2 subsets for Japanese, Korean, and Simplified Chinese. The target token ratio is 10\% English and 30\% each for Japanese/Korean/Chinese. 
To match this ratio without truncating text, we employ Weighted Dynamic sampling, which preserves natural document boundaries while converging to the target distribution.

The method uses an adaptive weight adjustment with exponential correction based
on distribution deviation:
\begin{align}
w_l^{t+1} &= w_l^t \cdot f(\delta_l) \\
f(\delta_l) &= \begin{cases}
1 + \alpha \cdot (e^{|\delta_l| \cdot \beta} - 1) & \text{if } \delta_l < 0 \\
\frac{1}{1 + \alpha \cdot (e^{|\delta_l| \cdot \beta} - 1)} & \text{if }
    \delta_l > 0
\end{cases} \\
\delta_l &= \frac{\text{actual}_l - \text{target}_l}{\text{target}_l}
\end{align}
where $w_l^t$ is the weight for language $l$, $\delta_l$ is the relative
deviation, and $\alpha, \beta$ are hyperparameters controlling adjustment
aggressiveness. Alternative sampling methods were evaluated and rejected; see
Appendix~\ref{app:sampling-ablations} for details.

\subsection{Optimization and Compute}
We train with AdamW ($\beta_1=0.9$, $\beta_2=0.95$, weight decay 0.1) and a
cosine learning-rate schedule with peak learning rate $3\times 10^{-4}$ and
2{,}000 warmup steps. The global batch processes 5.64M tokens per step across 8
GPUs. Training runs for 14{,}189 steps (80B tokens; one epoch over the sampled
stream), and we save checkpoints every 20 steps (112.8M tokens). For tractable
checkpoint-wise evaluation we subsample the saved checkpoints, yielding the 420
evaluated checkpoints reported in the results. Compute budget details are in
Appendix \ref{app:arch-budget}.
\section{Results}

Two evaluation protocols isolate UTF-8 generation capability from general
language modeling. Level 0 tests completion of valid UTF-8 byte sequences
without contextual cues; Level 1 tests retrieval of correct byte sequences when
semantic context constrains the target character. Results span 420 training
checkpoints from step 20 to step 14,189.

\subsection{Level 0: Context-Free Evaluation}

Level 0 evaluation tests the model's ability to generate valid UTF-8
continuations given only a partial byte sequence prefix, without any linguistic
context to guide its predictions. While this setting is relatively rare in
naturalistic text generation---rare characters seldom appear at the absolute
beginning of a sequence without any preceding context---it isolates whether
the model has internalized UTF-8 structural constraints independent of semantic
knowledge.

We evaluated the model using 2-byte prefixes, which we found to be more
diagnostic than single-byte prefixes. Single-byte prefixes provide insufficient
constraint, as the model can satisfy UTF-8 validity requirements through
multiple valid continuation patterns. With 2-byte prefixes, the model must
correctly identify whether the sequence requires additional continuation bytes
and, if so, generate bytes matching the required \texttt{10xxxxxx} pattern.

\subsubsection{Frequency and Validity}

\begin{figure}[t]
    
    \includegraphics[width=\linewidth]{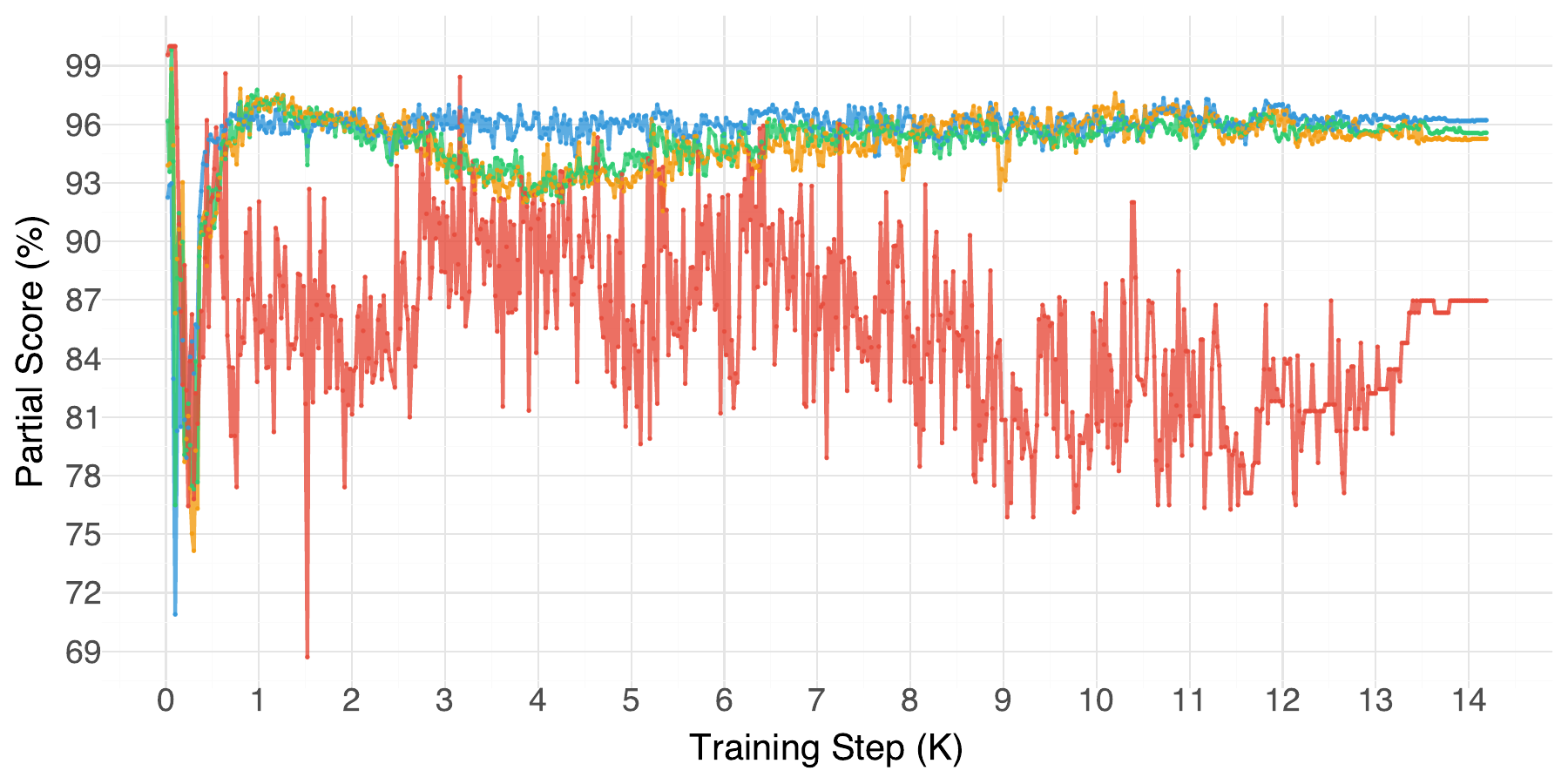}
    \caption{Per-tier plots for partial-credit validity. Common (blue), Uncommon (green), Rare (orange), Unseen (red).}
    \label{fig:per_tier}
\end{figure}

\begin{figure*}[t]
    \centering

    \begin{subfigure}[t]{0.48\textwidth}
        \centering
        \includegraphics[width=\linewidth]{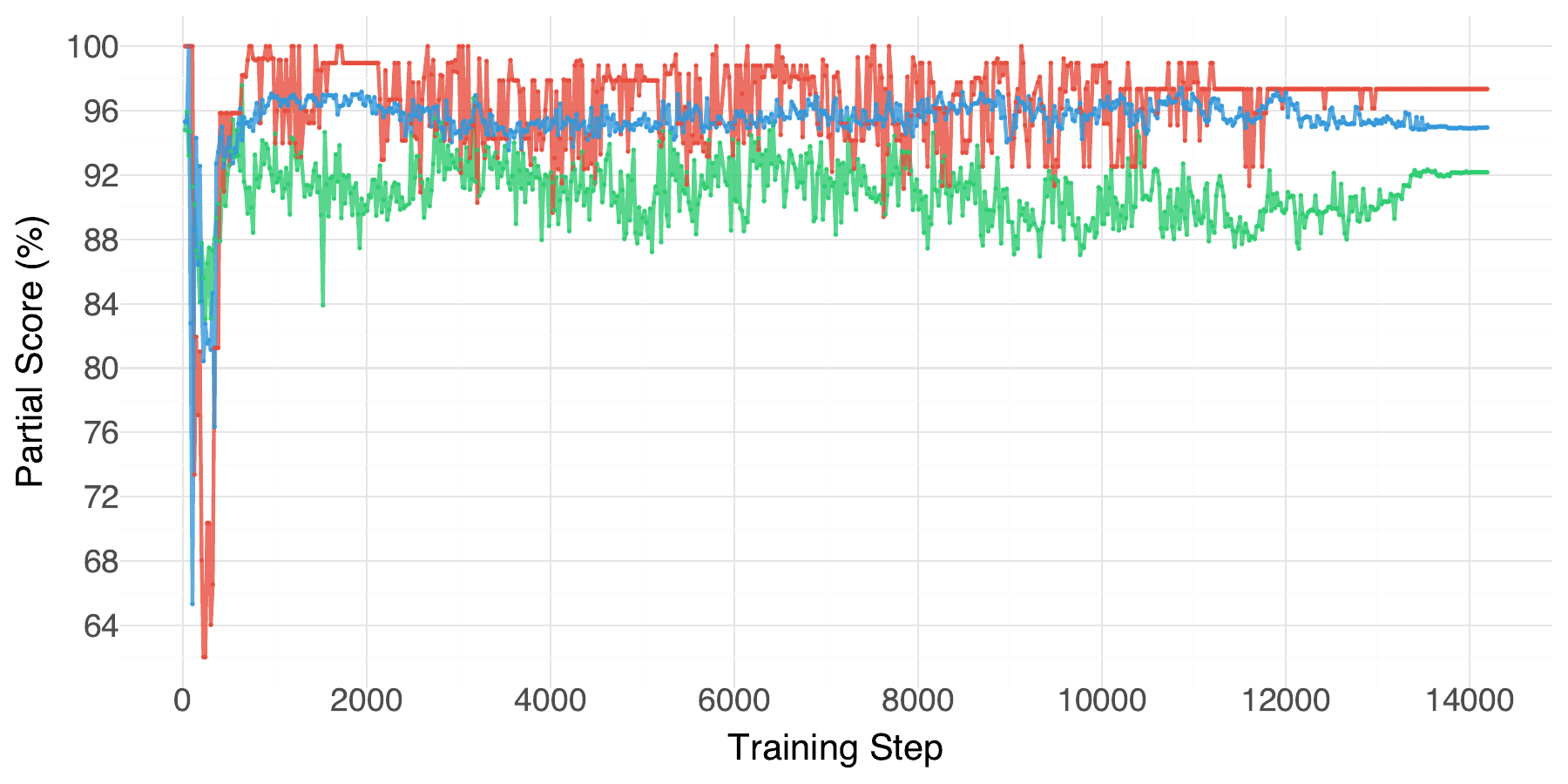}
        \caption{L0: Partial-Credit Validity}
    \end{subfigure}
    \begin{subfigure}[t]{0.48\textwidth}
        \centering
        \includegraphics[width=\linewidth]{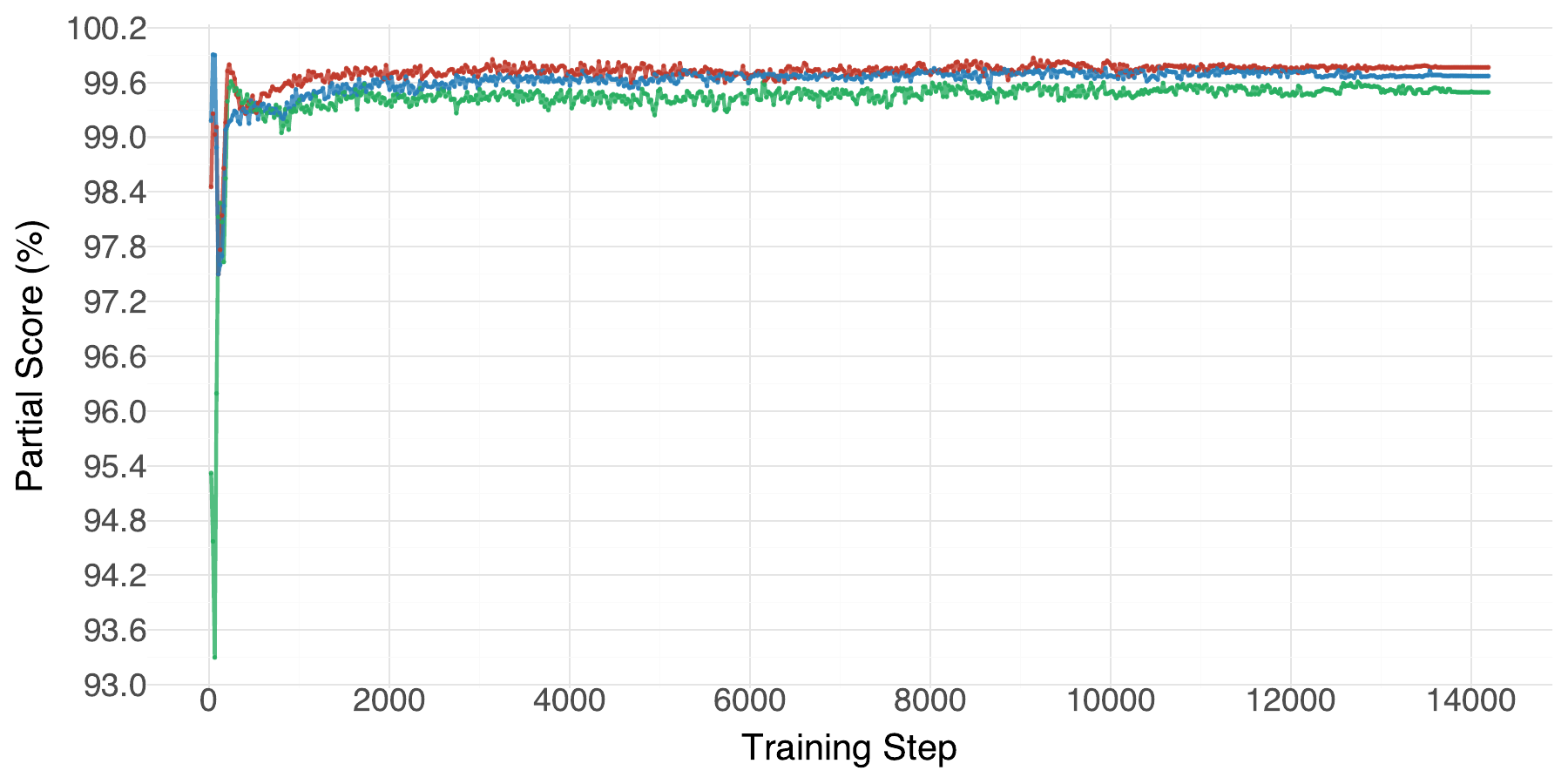}
        \caption{L1: Partial-Credit Validity}
    \end{subfigure}

    \begin{subfigure}[t]{0.48\textwidth}
        \centering
        \includegraphics[width=\linewidth]{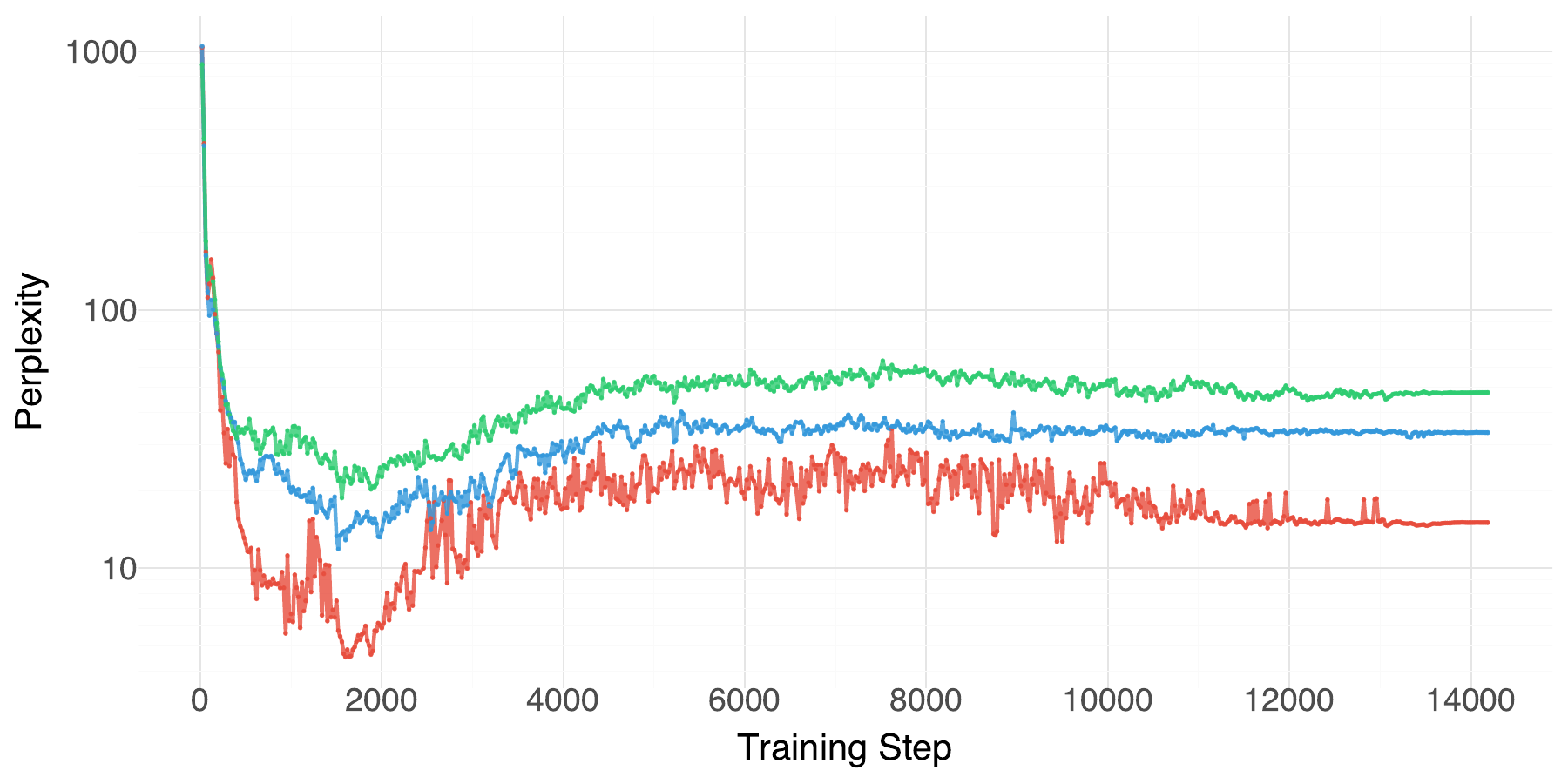}
        \caption{L0: Perplexity}
    \end{subfigure}
    \begin{subfigure}[t]{0.48\textwidth}
        \centering
        \includegraphics[width=\linewidth]{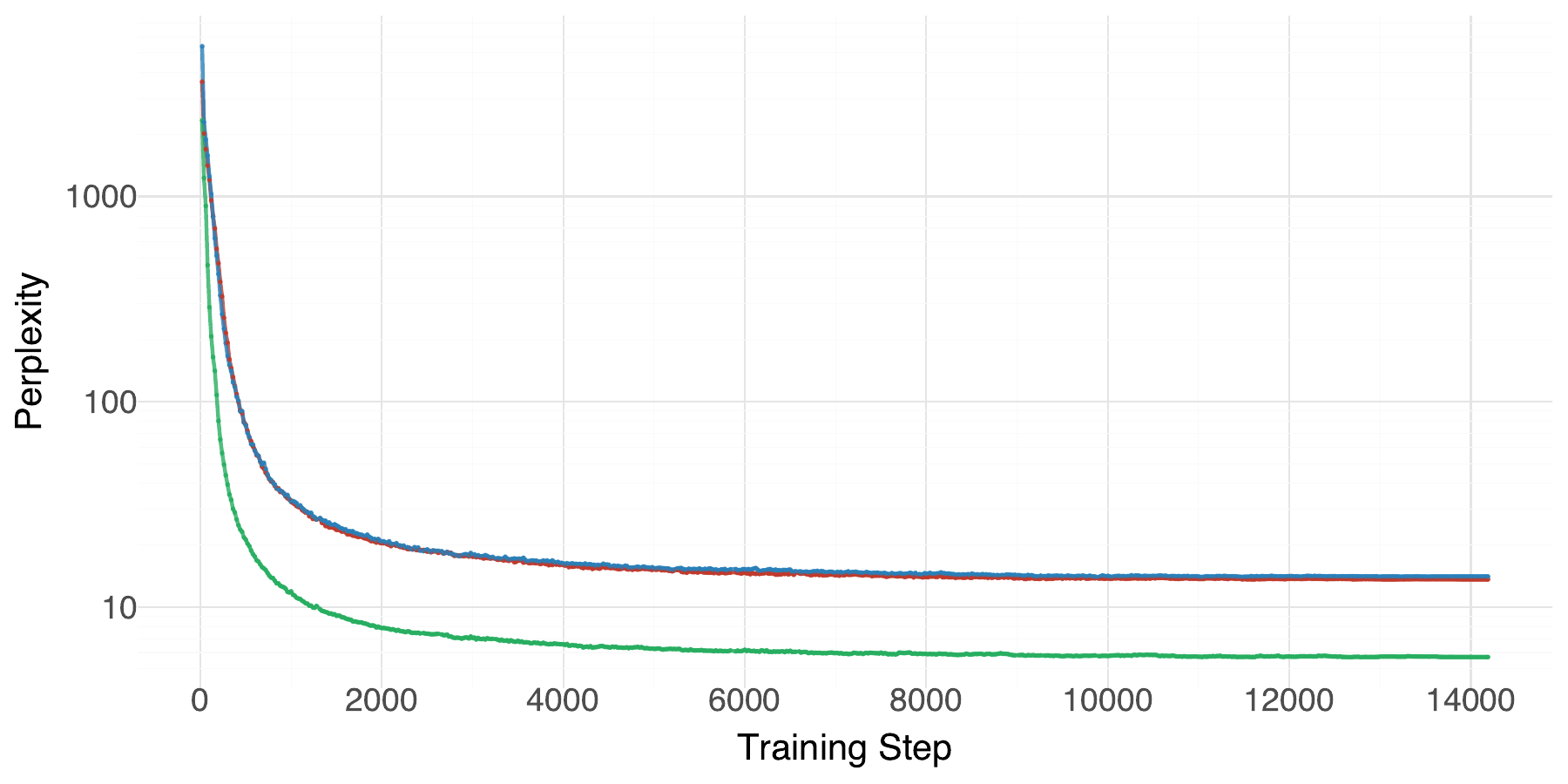}
        \caption{L1: Perplexity}
    \end{subfigure}

    \caption{Side-by-side comparison of learning dynamics. The \textbf{left column} shows the baseline (L0) and the \textbf{right column} shows the context-guided setting (L1). Results of Chinese, Japanese, and Korean are plotted in green, red, and blue, respectively. Note how the partial credit validity (top row) stabilize significantly faster in the L1 setting.}
    \label{fig:compare_grid}
\end{figure*}

Character frequency and structural validity exhibit a clear relationship in Figure~\ref{fig:per_tier}.
At the final checkpoint (step 14,189, corresponding to 80B tokens), the model achieved its highest partial-credit validity rates on the
\textit{Common} tier (96.21\%), followed by the \textit{Uncommon} tier (95.57\%), with the
\textit{Rare} tier close behind at 95.26\%. The \textit{Unseen} tier, containing characters
never observed during training, achieved 86.97\% partial-credit validity.
We use $V_{\text{partial}}$ as a learning-dynamics diagnostic throughout the
main results; strict validity, which most deployed applications require,
remains substantially lower (cf.\ Table~\ref{tab:cross-model}).

This pattern aligns with the intuition that more frequent exposure during training
leads to better internalization of byte-level structure. Characters in the Common
tier appear more often across diverse contexts, providing the model with more
opportunities to learn the correspondence between semantic content and UTF-8 byte
patterns. The monotonic decrease in validity from Common to Unseen tiers suggests
that byte-sequence generation capability is fundamentally tied to training
frequency, even when the characters themselves are out-of-vocabulary for the
subword tokenizer.

The Unseen tier's 86.97\% partial-credit validity rate, accompanied by the highest perplexity
among all tiers, demonstrates meaningful zero-shot generalization to novel
codepoints. The Unseen tier consists entirely of 4-byte UTF-8
characters (CJK Unified Ideographs Extension B and beyond, codepoints above
U+10000), whereas the Common, Uncommon, and Rare tiers contain 3-byte characters
from the Basic Multilingual Plane. This means the Unseen tier tests
generalization to a different byte-pattern family: sequences beginning with
\texttt{11110xxx} (4-byte) rather than \texttt{1110xxxx} (3-byte). The 86.97\%
partial-credit validity rate thus reflects the model's ability to generalize UTF-8 structural
rules across byte-length boundaries, despite never having encountered these
specific characters during training. Analysis of generated tokens indicates that
the model often correctly identifies the need for a 4-byte sequence but struggles
to select valid continuation bytes for start-byte combinations it has never
encountered. This partial generalization suggests that UTF-8 structural learning
involves both pattern-specific memorization from training exposure and limited
abstract rule induction.

\subsubsection{Byte-Length vs.\ Frequency}
\label{sec:unseen-3byte-control}

The Unseen tier in the main Level~0 set is dominated by 4-byte CJK Extension B
characters, which confounds two factors: novelty (the model has never seen the
character) and byte-length class (the model has seen few 4-byte sequences
overall). To separate them, we constructed a control set of 139 unseen 3-byte
CJK ideographs (exhaustive over the unseen 3-byte BMP characters in $U\setminus
K$) and evaluated 299 checkpoints under the same protocol. Table~\ref{tab:3byte-control}
reports partial-credit validity at prefix lengths 1--3 for the 4-byte Unseen
characters and the 3-byte control.

\begin{table}[t]
\centering
\small
\caption{Partial-credit validity (\%) on unseen 4-byte vs.\ 3-byte characters
across prefix lengths; prefix=3 crossover is structural.}
\begin{tabular}{@{}lrrr@{}}
\toprule
& Prefix=1 & Prefix=2 & Prefix=3 \\
\midrule
Unseen 4-byte         & 0.0 & 39.5 & 60.5 \\
Unseen 3-byte (control) & 48.2 & 51.1 & 19.4 \\
\bottomrule
\end{tabular}
\label{tab:3byte-control}
\end{table}

At prefix=1 the gap is 48.2~pp: the model continues 3-byte sequences from the
familiar \texttt{1110xxxx} lead byte at 48.2\% validity, but produces 0\%
valid UTF-8 from the novel \texttt{11110xxx} lead. Inspecting outputs at
prefix=1 for 4-byte targets, all 43 samples emit the identical bytes
\texttt{F0 9F 92 95} (U+1F495), the sole 4-byte character the model encountered
during training. The model has learned the 4-byte lead structure from this single
example but collapses to one template; it never falls back to a 3-byte lead.

Within the 3-byte class, frequency has little effect; final-checkpoint partial-credit validity ranges only from 
0.878 to 0.894 across Common,
Uncommon, Rare, and Unseen-3B groups.
The clear outlier is 4-byte Unseen (0.840).
This indicates that byte-length dominates failures at this scale.

\subsubsection{Relationship with Language Modeling}

Figure~\ref{fig:compare_grid} compares the evolution of partial-credit validity during training against that of perplexity and log-likelihood differences.
We observe that while perplexity quickly converges to a low level, partial-credit validity converges more slowly:
For partial-credit validity, convergence is observed after 740 training steps, while that for perplexity is observed as early as 380 training steps.
For log-likelihood differences, the tendency is similar to that of the perplexity.
We thus conclude that partial-credit validity does not correlate directly with either perplexity or log-likelihood relative to the gold standard. 
This suggests that our proposed metric captures a distinct aspect of a language model's ability to generate correct byte sequences, making it a useful and important complement to existing measures.

\subsection{Level 1: Context-Guided Evaluation}

Level 1 evaluation mirrors typical language model usage, where the model
generates text within semantic and syntactic context. Preceding words and
phrases constrain plausible continuations, enabling the model to retrieve
correct byte sequences by leveraging learned associations between concepts and
their byte-level representations. Most practical generation scenarios provide
such contextual cues.

\subsubsection{Structural and Semantic Evaluation}
\begin{figure}[t]
\includegraphics[width=\linewidth]{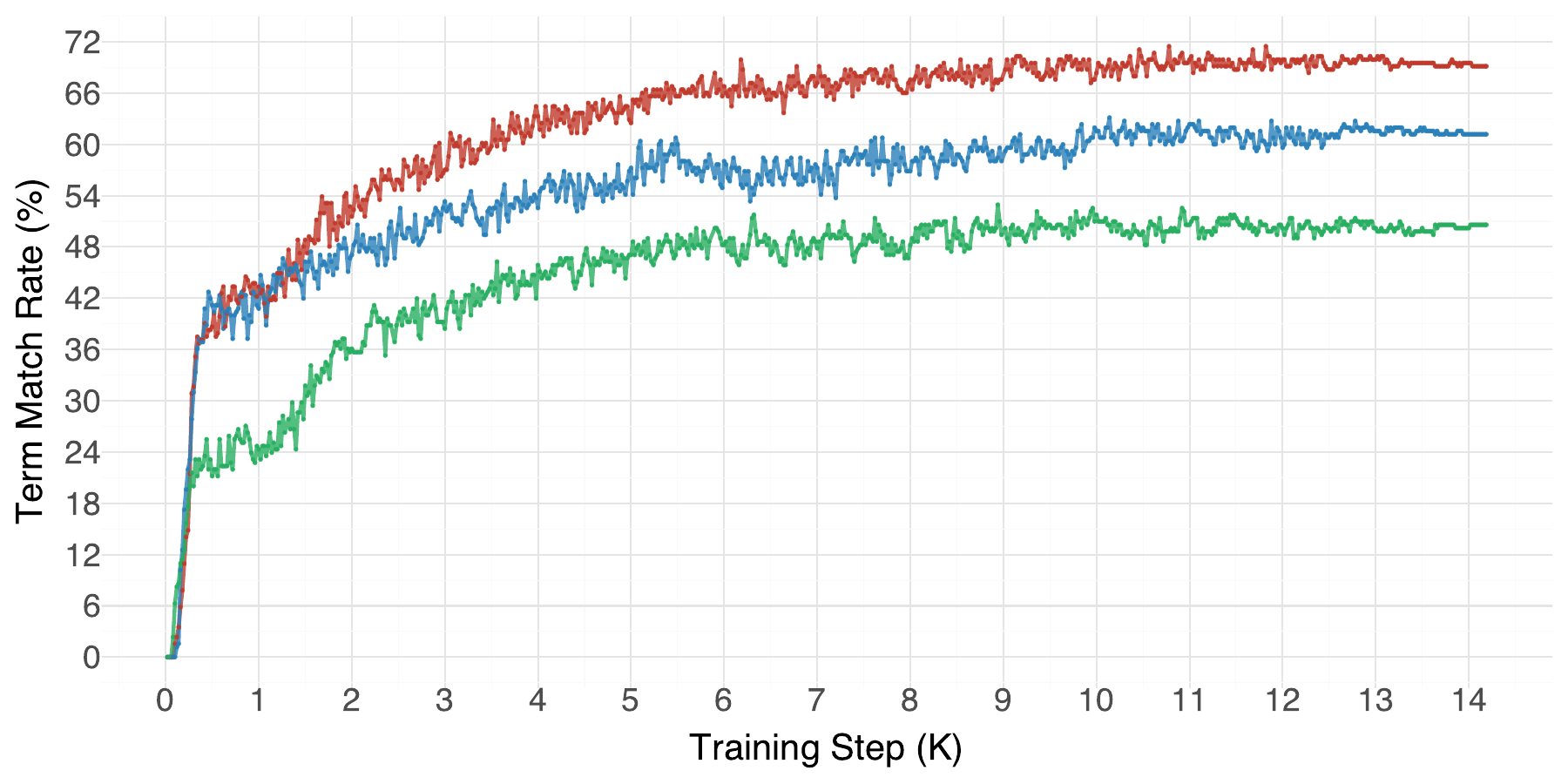}
\caption{Term match rate. Results of Chinese, Japanese, and Korean are plotted in green, red, and blue, respectively.}
\label{fig:term_match}
\end{figure}

Structural validity and semantic correctness diverge. As in Figure~\ref{fig:compare_grid}, the model achieves high
partial-credit validity in the context-guided setting, but Term Match Rate, i.e., whether it generates the specific correct character, 
reached 60.30\% (Figure~\ref{fig:term_match}).  The model
has mastered UTF-8 encoding \textit{mechanics} (generating valid byte sequences)
but struggles with \textit{semantics} (mapping context to the correct character).

The model often generates structurally valid characters that are semantically or
phonetically related to the target but not exact matches. When prompted with
context suggesting a particular kanji, the model may produce a different kanji
with similar radical structure or meaning. Byte-level syntax and byte-level semantics are distinct capabilities, with
the latter requiring more training.

The diagnostic $\Delta_{LL}$ (Eq.~\ref{eq:delta-ll}) splits the semantic
failures further. Among Level~1 failures where $\Delta_{LL}>0$ at the final
checkpoint (51 cases), the model assigns higher teacher-forced likelihood to
the gold continuation but greedy decoding emits a different byte. In 100\% of
these cases the emitted byte is a UTF-8 continuation byte in
\texttt{0x80}--\texttt{0xBF}, and for each lead byte the emission is a single
mode: \texttt{0xE3}~$\to$~\texttt{0x80} (CJK Symbols), \texttt{0xEC}~$\to$~\texttt{0x97}
(Common Hangul), \texttt{0xF0}~$\to$~\texttt{0x9F} (Emoji range). The failure
is mode collapse in $P(\text{byte}_2 \mid \text{byte}_1)$: the model has learned
the marginal argmax for each lead byte but does not condition on the target
character. Because $\Delta_{LL}>0$, beam search or temperature sampling may
recover these cases without retraining; verifying this empirically requires a
decoding ablation we leave to future work. Appendix~\ref{app:distractor}
reports the full distractor distribution.

\subsubsection{Context Helps Structural Learning}

Comparing Level 1 with Level 0 on the Common tier shows faster partial-credit validity
convergence in the context-guided setting (Figure~\ref{fig:compare_grid}). 
The model achieves reliable UTF-8
generation earlier when semantic context is provided, suggesting contextual
associations facilitate byte sequence retrieval before the model has fully
internalized UTF-8 structural rules in isolation.

Cross-language patterns in Level 1 parallel Level 0: Japanese characters achieve
high validity earlier than Korean and Chinese. The gap between languages is
narrower in Level 1, indicating that context partially compensates for sparser
byte-pattern exposure in larger character inventories.

\subsubsection{Relationship with Language Modeling}

Figure~\ref{fig:compare_grid} compares the evolution of partial-credit validity during training against that of perplexity and log-likelihood differences.
The model generates a single token given a 2-byte prefix.
Unlike Level 0 evaluation, we observe that in Level 1, partial-credit validity converges faster than perplexity and log-likelihood differences.
This suggests that generating byte sequences with contextual guidance (Level 1) is easier than context-free generation (Level 0).
Nevertheless, the observation that partial-credit validity converges faster than perplexity further underscores the need for a metric beyond perplexity to evaluate a model's ability to generate valid byte sequences, highlighting the importance of our proposed metrics.

The finding that partial-credit validity converges faster than perplexity in Level
1, while the opposite holds in Level 0, further underscores that these metrics
capture fundamentally different aspects of model capability. Perplexity measures
the model's uncertainty over the full continuation distribution, whereas UTF-8
validity measures only whether the generated sequence satisfies structural
encoding constraints. The divergent convergence patterns across evaluation
levels demonstrate that neither metric subsumes the other, validating the need
for dedicated partial-credit validity evaluation in byte-level language models.

\section{Related Work}
Byte-level tokenization eliminates vocabulary bottlenecks and handles the full
diversity of Unicode characters, particularly for morphologically rich and
logographic languages where subword tokenization faces inherent limitations.
\citet{xue-etal-2022-byt5} introduced ByT5, demonstrating that byte-level models
could match the performance of token-based models while offering improved noise
robustness. Their work showed that operating directly on UTF-8 bytes eliminates
the need for language-specific preprocessing and handles any Unicode input
without information loss. More recently, \citet{custom-pagnoni2024byte}
presented the Byte Latent Transformer (BLT), achieving comparable performance to
Llama 3 while using 50\% fewer inference FLOPs, suggesting byte-level
architectures may offer computational advantages at scale.

The challenges of Unicode processing in LLMs have been documented across
multiple dimensions. \citet{rust-etal-2021-good} demonstrated that
morphologically rich languages require significantly more tokens than English
for equivalent semantic content, creating systematic biases in multilingual
models. The long-tail distribution of characters exacerbates this
problem. For example, \cite{singh-etal-2024-aya} found that low-resource languages suffer
from poor tokenization efficiency, leading to degraded performance that
discourages their use in training data, creating a vicious cycle.

Recent security research has revealed how tokenization vulnerabilities can be
exploited. \citet{custom-geh2024adversarial} discovered that LLMs retain
semantic understanding of non-canonical tokenizations despite never encountering
them during training, enabling attackers to bypass safety filters through
alternative word segmentations. The ``glitch token'' phenomenon, analyzed by
\citet{custom-li2024glitch} and further investigated by
\citet{land-bartolo-2024-fishing}, identified tokens like ``SolidGoldMagikarp''
that cause unpredictable model behavior. These tokens cluster in the embedding
space and result from insufficient training.

Prior art addresses invalid byte generation at decode time rather
than at training time. Constrained decoding masks tokens whose continuations
would violate a grammar or automaton, guaranteeing that the emitted sequence
stays in a target language~\citep{willard2023efficient,koo2024automata}.
\citet{cognetta-okazaki-2025-tokenization} formalize BPE and WordPiece
as finite-state transducers, enabling subword-aware pattern promotion that is
consistent with the tokenizer. Encoding UTF-8 validity in the same framework
removes structural failures by construction, since UTF-8 is a regular language
over bytes (\S~\ref{sec:utf8-dfa-metric}). Our work is complementary: we
measure when byte-level models acquire UTF-8 competence during training and
separate the structural failure mode addressed by constrained decoding from the
semantic failure mode (Term Match, Sec.~\ref{sec:term-match}) that it leaves
unchanged.

While \citet{custom-kaplan2020scaling} and \citet{custom-hoffmann2022training}
established power-law relationships between model scale and performance,
subsequent work has shown these relationships break down for multilingual and
Unicode-sensitive tasks. \citet{custom-pokharel2025scaling} demonstrated that
in zero-shot multilingual scenarios, model scale has minimal effect on
performance. This suggests that scaling alone cannot overcome the fundamental
vocabulary bottleneck created by tokenization, motivating our investigation
into the minimum scale required for reliable UTF-8 sequence generation.
Our findings through the UTF-8 expression can be applied to systems using its alternatives proposed in recent years~\citep{limisiewicz-etal-2024-myte,land2026bpe}. 
\section{Discussion}

Languages vary in UTF-8 validity learning rates. Japanese reaches reliable
validity earlier than Korean and Chinese, but two evaluation-set properties
contribute beyond character inventory size. The Japanese OOV set contains 36
characters (mostly Kana, since most Kana are vocabulary-covered) compared with
256 for Korean and 256 for Chinese, which biases the language-level average
upward. Kana also occupy a narrow Unicode range (U+3040--U+30FF) with regular
\texttt{E3~8x~xx} byte patterns, whereas Hangul (lead bytes \texttt{EA}--\texttt{ED})
and CJK ideographs (\texttt{E4}--\texttt{E9}) span a much larger byte-pattern
space. The faster Japanese convergence reflects these structural and sampling
properties, not a language-specific capability. The Unseen tier's 86.97\%
validity on 4-byte characters never encountered during training reflects
generalization across byte-length boundaries, with the byte-length dimension
identified as the dominant axis in our control experiment
(Sec.~\ref{sec:unseen-3byte-control}).

Our findings are not an argument against smaller models, which remain essential
for edge devices, real-time systems, and resource-constrained environments.
Smaller models serve as research tools, distillation targets, and domain-specific
solutions. The findings instead clarify training requirements for byte-level
tokenization and metrics to monitor for reliable UTF-8 generation.

\subsection{Cross-Model Validation}

To test whether the structural-semantic gap generalizes beyond our 355M baseline,
we evaluated 10 open-weight models from 5 families (1B--9B parameters) using the
same Level~0 and Level~1 protocol and evaluation data. Results are summarized in
Table~\ref{tab:cross-model}; checkpoint sources,
tokenizer handling, and decoding settings are reported in Appendix~\ref{app:cross-model-details}.

\begin{table}[t]
\centering
\small
\caption{Cross-model evaluation at Level~0/1. $V_p$: partial-credit
validity, $V_s$: strict validity, $M$: term match rate. All values (\%) at
generation step~5, averaged across languages and prefix lengths.}
\begin{tabular}{@{}llcccc@{}}
\toprule
\textbf{Family} & \textbf{Size} & \textbf{$V_p$} & \textbf{$V_s$} & \textbf{Gap} & \textbf{L1 $M$} \\
\midrule
Baseline & 0.4B & 93.5 & 39.8 & 53.6 & 47.8 \\
\midrule
Gemma-3 & 1B & 33.6 & 0.9 & 32.6 & 0.3 \\
Gemma-3 & 4B & 100 & 35.7 & 64.3 & 0.0 \\
\midrule
Llama-3.2 & 1B & 96.9 & 38.4 & 58.5 & 0.8 \\
Llama-3.2 & 3B & 97.8 & 49.0 & 48.8 & 1.0 \\
Llama-2 & 7B & 98.4 & 33.9 & 64.5 & 33.1 \\
\midrule
Mistral & 7B & 96.0 & 49.9 & 46.1 & 23.3 \\
\midrule
OLMo-2 & 1B & 93.0 & 33.5 & 59.5 & 7.0 \\
OLMo-2 & 7B & 93.5 & 28.0 & 65.4 & 14.3 \\
\midrule
Qwen-3.5 & 4B & 99.2 & 87.0 & 12.2 & 2.4 \\
Qwen-3.5 & 9B & 99.7 & 89.8 & 9.9 & 0.5 \\
\bottomrule
\end{tabular}
\label{tab:cross-model}
\end{table}

The $V_p$--$V_s$ gap persists across all models and scales tested (9.9--65.4\,pp).
Gap magnitude is stable within families, where OLMo-2 shows 60--65\,pp gaps at both
1B and 7B; Qwen-3.5 shows 10--12\,pp at both 4B and 9B, but varies substantially
across families.
$V_s$ generally improves with scale (Llama-3.2: +10.6\,pp from 1B to 3B;
Qwen-3.5: +2.8\,pp from 4B to 9B), but Level~1 term match does not improve
consistently.
This confirms that structural validity and semantic correctness are distinct:
larger models produce more valid byte sequences, but often not the
\emph{correct} character.

We cannot measure convergence rates on the open models without checkpoint-level
sweeps, but cross-model evidence is consistent with the convergence-lag
hypothesis from our baseline. If $V_s$ caught up to $V_p$ given enough training,
we would expect smaller gaps in larger models that have seen more tokens, and
the data does not show this trend.

Tokenizer design is a stronger predictor of semantic byte completion than model
size. SentencePiece byte-fallback models (baseline at 47.8\%, Llama-2 at 33.1\%,
Mistral at 23.3\%) achieve higher term match than GPT-2-style BPE models with
larger vocabularies (Qwen-3.5~9B at 0.5\%, Gemma-3~4B at 0.0\%). Small
vocabularies force frequent byte-fallback during training, giving the model
more practice at byte-level generation.

\subsection{Limitations and Future Work}

Our training dynamics analysis (the convergence gap between perplexity and
validity) is conducted at a single 355M-parameter training run; ``scale-dependent''
here refers to training-token scale under this setup rather than a scaling law.
The cross-model evaluation confirms that the structural-semantic
gap exists at larger scales, but we do not have checkpoint-level sweeps
for open models to measure convergence \emph{rates} at those scales.
We also emphasize that $V_{\text{partial}}$ is a diagnostic; deployed applications
typically require fully decodable strings, for which $V_{\text{strict}}$ is the
operational metric.

Our Level~1 contexts are generated by Gemini~3~Pro and filtered with formal
criteria (language identification, uniqueness, length); we do not apply human
semantic validation. Context naturalness, target-character appropriateness, and
synthetic-generator bias can therefore affect Term Match measurements, and the
synthetic distribution may differ from naturally occurring text.

Our experiments focus on East Asian languages (Chinese, Japanese, Korean) plus
English. These languages were chosen for their diverse character sets and
multi-byte encoding requirements, but findings may not replicate to other
languages. The evaluation framework itself is
language-agnostic and applicable to any script encoded in UTF-8.

Constrained decoding methods~\citep{koo2024automata,willard2023efficient,cognetta-okazaki-2025-tokenization}
can guarantee $V_{\text{strict}} = 1.0$ by masking
structurally invalid tokens at each generation step. However, our distractor
analysis (Appendix~\ref{app:distractor}) shows that in 100\% of $\Delta_{LL} > 0$
failure cases, the model already produces structurally valid continuations: the
failures are semantic, not structural. Constrained decoding addresses
structural failures, but not semantic ones. Since $\Delta_{LL}>0$ implies the
model assigns higher probability to the gold completion under teacher forcing,
beam search or temperature sampling may recover the corresponding cases
without retraining, though we leave a controlled decoding ablation to future
work.
The constrained-decoding intervention rate may serve as a training diagnostic, warranting further study.

A related question is whether the validity lag is specific to byte-level
generation or a special case of slower learning on rare tokens. The two-fold
lag we report compares byte-level validity against overall perplexity, which
is dominated by common tokens, so part of the gap is the standard long-tail
versus common-case gap. Two observations argue for distinct dynamics. First,
the failure mode is concentrated mode collapse rather than distributed
uncertainty: in 100\% of $\Delta_{LL}>0$ cases, each lead byte maps to a single
continuation byte (Appendix~\ref{app:distractor}). Generic rare-token failures
show a spread of plausible alternatives. Second, tokenizer setup overrides
scale in the cross-model evaluation: SentencePiece byte-fallback models with
small vocabularies outperform much larger models with large vocabularies on
term match. Cleanly separating byte-level dynamics from rare-subword dynamics
would require two ablations we leave for future work: tracking per-cohort
cross-entropy on subword tokens versus byte-fallback sequences sampled at
matched training frequencies, and training two otherwise-identical models that
differ only in CJK vocabulary coverage.

Architectural changes such as explicit byte-position encodings or
hierarchical representations may accelerate UTF-8 learning. Context-rich
generation achieves reliable UTF-8 output at lower training scales than
context-sparse generation, though structural validity does not guarantee
semantic correctness, and the tradeoff needs further investigation.

\section{Conclusion}

UTF-8 validity is a distinct capability that emerges at different rates than
standard language modeling metrics. In experiments with a 355M parameter model
evaluated across 420 checkpoints, UTF-8 validity convergence lagged perplexity
convergence, perplexity stabilizes after approximately 2.1B training tokens,
while UTF-8 validity requires roughly 4.2B tokens, a two-fold difference.
Practitioners deploying byte-level models should note that a model appearing
well-trained by perplexity standards may still produce invalid UTF-8 sequences.

Character frequency correlates with structural validity (Common: 96.21\%,
Uncommon: 95.57\%, Rare: 95.26\%), but a control experiment indicates that
byte-length exposure is the dominant axis of failure at this scale, with
frequency contributing only modestly within a fixed byte-length class
(Sec.~\ref{sec:unseen-3byte-control}).
The gap between structural validity and semantic
correctness remains stark: despite moderate validity rates, Term Match Rate
reached 60.30\%. 
Languages with larger character inventories (e.g., Korean and Chinese) need more training exposure than Japanese to reach comparable validity.

Context-guided evaluation shows that semantic context accelerates structural
validity convergence, though semantic correctness remains low even with context.
Training data exposure should exceed perplexity convergence thresholds, and
UTF-8 validity should be monitored as a distinct metric.

\clearpage

\section*{Acknowledgements}
This work was supported by JSPS KAKENHI Grant Number 25H01137, 
the ``R\&D Hub Aimed at Ensuring Transparency and Reliability of Generative AI Models'' project of the Ministry of Education, Culture, Sports, Science and
Technology, 
and JST K Program Japan Grant Number JPMJKP24C3.

\section*{Impact Statement}
This paper presents work whose goal is to advance the field of Machine Learning. There are many potential societal consequences of our work, none which we feel must be specifically highlighted here. 

\bibliographystyle{icml2026}

\newpage
\appendix
\onecolumn
\section{Ethical Considerations}

This research investigates potential vulnerabilities in language model
tokenization that could be exploited for adversarial purposes. However, we
believe the benefits of understanding these limitations outweigh the risks. Our
work aims to characterize vulnerabilities so they can be addressed in future
model architectures, ultimately improving the robustness of deployed systems. We
do not introduce new attack methodologies but rather study the scaling
properties of known tokenization issues. All adversarial sequences tested are
synthetically generated rather than optimized for maximum harm. Our research
provides concrete guidance on training requirements for reliable byte-level
generation, helping practitioners make informed decisions about when byte-level
tokenization is viable for their applications.

We have deliberately avoided testing our methods on production models or
developing tools that could facilitate malicious use. All experiments are
conducted on models we train ourselves, ensuring no impact on deployed systems.
We find no significant ethical risks associated with this work beyond the
general considerations of academic research.

\section{Model Architecture and Compute Budget}
\label{app:arch-budget}

The GPT-2 variant model used for our work employs several standard modernization choices
for training stability and efficiency. We replace LayerNorm with RMSNorm \citep{custom-zhang2019root},
use Rotary Position Embeddings (RoPE) \citep{su-etal-2024-roformer}, and adopt
Grouped-Query Attention (GQA) \citep{custom-ainslie2023gqa} with a reduced number
of key/value heads to lower memory usage (similar in spirit to recent
implementations such as Gemma 2 \citep{custom-team2024gemma}). The feed-forward
blocks use a gated MLP (GeGLU variant) \citep{custom-shazeer2020glu}. We further
apply query scaling by $d_{\text{head}}^{-0.5}$ and embedding scaling by
$\sqrt{d_{\text{hidden}}}$. 

Training was conducted on a single node with 8 Nvidia B200 GPUs for one epoch over
76 hours. Evaluation used various single-accelerator setups across A6000s,
RTX PRO 6000s, and an M4 Pro Mac Mini.

\section{Training Corpus Sampling Method Ablations}
\label{app:sampling-ablations}

\begin{figure*}[t]
    \centering
    \includegraphics[width=0.9\linewidth]{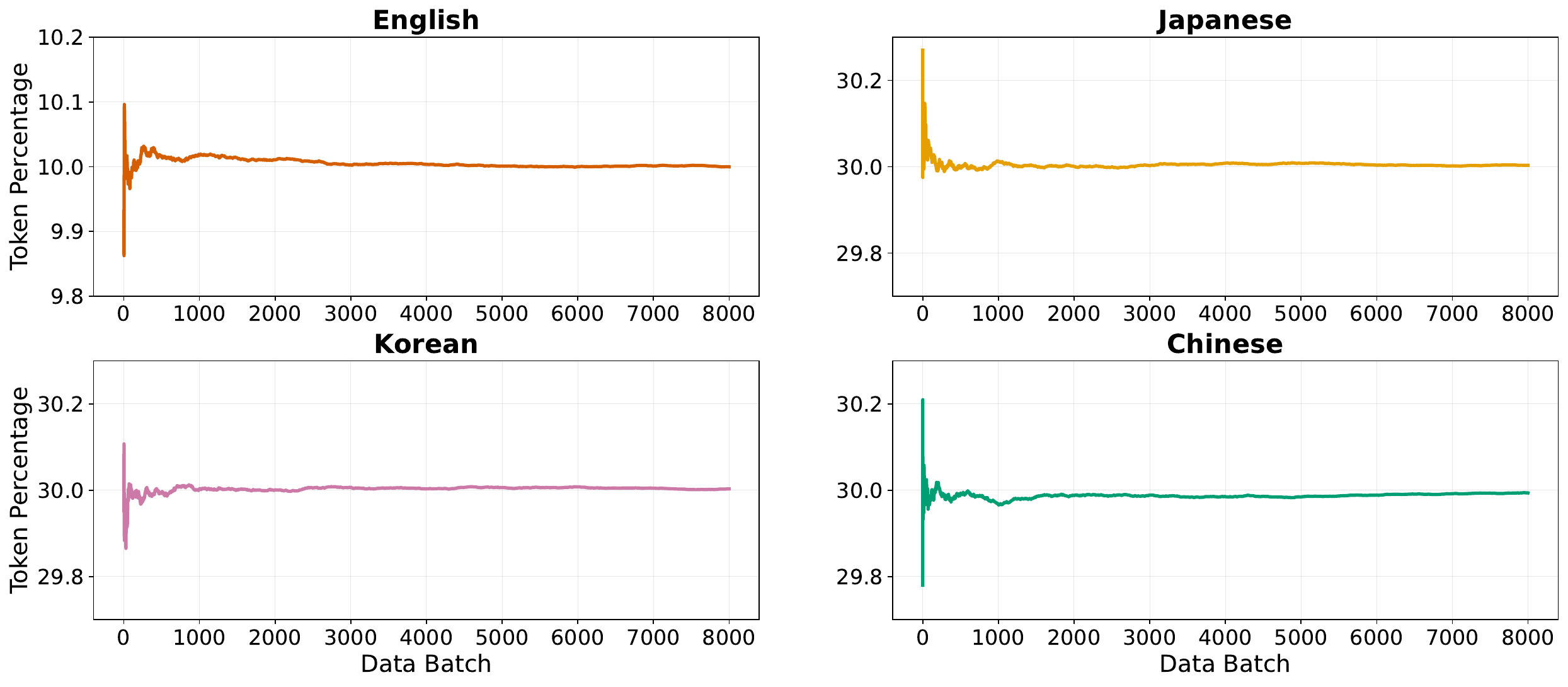}
    \caption{Token distribution convergence during training data construction.
    The adaptive weight adjustment algorithm dynamically modifies language
    sampling probabilities to achieve the target distribution while preserving
    natural sentence boundaries.}
    \label{fig:tokenization-balance}
\end{figure*}

The main paper describes Weighted Dynamic sampling as the chosen method for
training corpus construction. This appendix documents the alternative methods
that were evaluated and rejected. Weighted Dynamic sampling was chosen because
alternative methods either truncate sentences mid-thought (Strict Quota) or
produce unnatural cross-lingual mixing within batches (Buffer-Balanced).

\subsection{Strict Quota}

The Strict Quota method enforces exact token distribution by maintaining running
quotas for each language, defined as follows:

\begin{align}
Q_l^{t+1} &= \max(0, Q_l^t - |d_l|) \\
P(l|Q) &= \begin{cases}
1 & \text{if } Q_l > 0 \text{ and } Q_l = \max_k Q_k \\
0 & \text{otherwise}
\end{cases}
\end{align}

where $Q_l^t$ is the remaining quota for language $l$ at time $t$, and $|d_l|$ is
the token count of document $d$ from language $l$. This method was rejected because
it frequently truncates sentences mid-thought when quotas are exhausted, producing
training examples that do not reflect natural language boundaries and potentially
teaching the model to generate incomplete utterances. 

\subsection{Buffer-Balanced}

The Buffer-Balanced method maintains separate token buffers for each language and
extracts proportionally from each buffer, defined as follows:

\begin{align}
B_l^{t+1} &= B_l^t \cup \text{tokens}(d_l) - E_l^t \\
E_l^t &= \text{extract}(B_l^t, n_l) \\
n_l &= \lfloor p_l \cdot S \rfloor
\end{align}

where $B_l^t$ is the buffer for language $l$, $E_l^t$ is the extracted tokens,
$p_l$ is the target proportion, and $S$ is the block size.

\subsection{Qualitative Comparison}

Table~\ref{tab:sampling-examples} shows detokenized text from 256-token training
batches for each method.

\begin{table*}[t]
\centering
\caption{Example 256-token training batches from each sampling method. Strict Quota enforces exact distribution but truncates mid-sentence. Buffer-Balanced mixes languages unnaturally within blocks. Weighted Dynamic preserves sentence boundaries at the cost of distribution precision.}
\small
\begin{tabular}{p{2.2cm}p{13cm}}
\toprule
\textbf{Method} & \textbf{Example Batch Content} \\
\midrule
Strict Quota &
\textbf{[EN: 26 tok]} The quick brown fox jumps over the lazy dog while\ldots \newline
\textbf{[KO: 77 tok]} \begin{CJK}{UTF8}{mj}오늘은 날씨가 매우 좋아서 공원에 산책을 나갔습니다. 봄꽃들이 활짝 피어있고 새들이 노래를 부르고 있었어요. 많은 사람들이 가족과 함께 즐거운 시간을\end{CJK} \newline
\textbf{[JA: 77 tok]} \begin{CJK}{UTF8}{min}今日は天気がとても良かったので、公園に散歩に行きました。春の花が満開で、鳥たちが歌っていました。多くの人々が家族と一緒に楽しい時間を過ごしていました。でも急に雨が降り始めて、みんな慌てて\end{CJK} \newline
\textbf{[ZH: 76 tok]} \begin{CJK}{UTF8}{gbsn}今天天气非常好，所以我去公园散步了。春天的花朵盛开着，鸟儿在唱歌。许多人和家人一起度过愉快的时光。但是突然开始下雨了，大家都慌慌张张地跑回家。我也赶紧找了个地方避雨，等雨停了才\end{CJK} \newline
\textit{Text cuts off mid-sentence when quota is reached.} \\
\midrule
Buffer-Balanced &
The implementation follows a simple \begin{CJK}{UTF8}{mj}프로그래밍 패턴을 사용하여\end{CJK} pattern where each \begin{CJK}{UTF8}{min}コンポーネントは独立して\end{CJK} component operates \begin{CJK}{UTF8}{gbsn}独立运行并且可以\end{CJK} independently and can be easily replaced. \begin{CJK}{UTF8}{mj}이러한 방식은 유지보수를 쉽게 만들고\end{CJK} \begin{CJK}{UTF8}{min}メンテナンスが簡単になり、\end{CJK} \begin{CJK}{UTF8}{gbsn}系统的灵活性也大大提高了。\end{CJK} \newline
\textit{Unnatural language mixing within single training blocks.} \\
\midrule
Weighted Dynamic &
\textbf{[EN: 19 tok]} Machine learning has revolutionized how we process data. \newline
\textbf{[KO: 82 tok]} \begin{CJK}{UTF8}{mj}기계 학습은 우리가 데이터를 처리하는 방식에 혁명을 일으켰습니다. 특히 딥러닝 기술의 발전으로 이미지 인식, 자연어 처리, 음성 인식 등 다양한 분야에서 놀라운 성과를 거두고 있습니다.\end{CJK} \newline
\textbf{[JA: 71 tok]} \begin{CJK}{UTF8}{min}機械学習は私たちのデータ処理方法に革命をもたらしました。特にディープラーニング技術の発展により、画像認識、自然言語処理、音声認識など様々な分野で驚くべき成果を上げています。\end{CJK} \newline
\textbf{[ZH: 84 tok]} \begin{CJK}{UTF8}{gbsn}机器学习彻底改变了我们处理数据的方式。特别是深度学习技术的发展，在图像识别、自然语言处理、语音识别等各个领域都取得了令人惊叹的成果。这些技术正在改变我们的日常生活。\end{CJK} \newline
\textit{Natural sentence boundaries preserved; distribution less precise.} \\
\bottomrule
\end{tabular}
\label{tab:sampling-examples}
\end{table*}

\subsection{Quantitative Results}

The analysis reveals a fundamental trade-off between distribution precision and text
quality. Strict Quota achieves near-perfect distribution (0.44\% MAE) with varying
byte-fallback rates per language. Buffer-Balanced shows moderate distribution
accuracy (1.36\% MAE) with intermediate byte-fallback rates. Weighted Dynamic,
despite higher distribution error (2.70\% MAE), maintains the lowest byte-fallback
rates across all languages, particularly for CJK languages.

\definecolor{bestcolor}{rgb}{0.7, 0.9, 0.7}

\begin{table*}[t]
\centering
\caption{Performance metrics across 20 trials (100 blocks each)}
\begin{tabular}{lcccc}
\toprule
Method & MAE (\%) & RMSE (\%) & Time (ms) \\
\midrule
Strict Quota & 0.44 & 0.52 & 85.5 \\
Buffer-Balanced & 1.36 & 1.60 & 96.6 \\
Weighted Dynamic & 2.70 & 3.12 & 76.0 \\
\bottomrule
\end{tabular}
\end{table*}

\section{Language-Specific Accuracy and Byte-Fallback Statistics}

This appendix reports two per-language measurements taken on the training
corpus. The first quantifies how closely each sampling method matches the
target language proportions in the assembled corpus, complementing the
aggregate MAE/RMSE numbers in Appendix~\ref{app:sampling-ablations}. The
second quantifies how much of the resulting training signal arrives through
byte-fallback tokens rather than full subword tokens; this is the quantity
most directly relevant to byte-level learning dynamics studied in the main
paper.

Table~\ref{tab:lang-accuracy} reports the realized per-language token share
across 20 trials of each sampling method. Strict Quota achieves the lowest
deviation (as expected, since it enforces quotas by construction), at the
cost of mid-sentence truncations documented in
Appendix~\ref{app:sampling-ablations}. Buffer-Balanced and Weighted Dynamic
both keep deviations within a few percentage points of target; we adopt
Weighted Dynamic for the main training run because it preserves sentence
boundaries while still tracking the target distribution.

\begin{table*}[t]
\centering
\caption{Language-specific distribution accuracy. Values are the realized
percentage of training tokens in each language (mean $\pm$ s.d.\ across 20
trials of 100 blocks each).}
\label{tab:lang-accuracy}
\begin{tabular}{llrrr}
\toprule
Language & Target & Strict Quota & Buffer-Balanced & Weighted Dynamic \\
\midrule
EN & 10.0 & 10.01 ± 0.36 & 10.51 ± 2.68 & 10.57 ± 4.65 \\
KO & 30.0 & 29.69 ± 0.57 & 29.37 ± 1.50 & 28.80 ± 2.55 \\
JA & 30.0 & 30.18 ± 0.57 & 30.26 ± 1.29 & 29.98 ± 2.80 \\
ZH & 30.0 & 30.11 ± 0.56 & 29.85 ± 0.89 & 30.65 ± 3.26 \\
\bottomrule
\end{tabular}
\end{table*}

Table~\ref{tab:byte-fallback-rates} reports the byte-fallback rate per
language, i.e., the fraction of training tokens emitted as raw byte tokens
rather than subword tokens within each language partition. Chinese drives
most of the byte-fallback load: at 30\% of the corpus with a 56.1\%
byte-fallback rate, it contributes roughly 0.30 $\times$ 0.561 / 0.227
$\approx$ 74\% of all byte-fallback tokens seen during training. English is
effectively absent from the byte-fallback signal, and Japanese/Korean
contribute moderately. These per-language rates explain why the model has
markedly more exposure to 3-byte CJK ideograph patterns than to other
multi-byte structures and provide quantitative grounding for the
byte-length effects analyzed in Section~\ref{sec:unseen-3byte-control}.

\begin{table}[t]
\centering
\caption{Per-language byte-fallback rates measured on the trained corpus
under the Weighted Dynamic sampling configuration. The byte-fallback rate is
the fraction of training tokens emitted as byte-fallback tokens within each
language partition. Total tokens 45.47B; byte-fallback tokens 10.30B;
non-byte tokens 35.16B.}
\label{tab:byte-fallback-rates}
\begin{tabular}{lrr}
\toprule
Language & Corpus share & Byte-fallback rate \\
\midrule
English  & 10\% &  0.2\% \\
Japanese & 30\% & 11.1\% \\
Korean   & 30\% &  8.3\% \\
Chinese  & 30\% & 56.1\% \\
\midrule
Overall  & 100\% & 22.7\% \\
\bottomrule
\end{tabular}
\end{table}

\section{UTF-8 DFA Transition Details}
\label{app:utf8-dfa}

The UTF-8 validation DFA (Figure~\ref{fig:utf8-dfa} in the main text) implements
the full UTF-8 specification with explicit rejection of malformed sequences. The
automaton has eight states: $S_0$ (initial/accepting), $S_1$ (awaiting 2-byte
continuation), $S_2$ and $S_{2,1}$ (3-byte sequence states), $S_3$, $S_{3,1}$,
and $S_{3,2}$ (4-byte sequence states), and $S_{\text{err}}$ (error sink).

Transitions from $S_0$ are determined by the lead byte:
\begin{itemize}
\item \texttt{00-7F}: ASCII, self-loop to $S_0$
\item \texttt{C2-DF}: 2-byte lead, transition to $S_1$
\item \texttt{E0-EF}: 3-byte lead, transition to $S_2$
\item \texttt{F0-F4}: 4-byte lead, transition to $S_3$
\item \texttt{80-BF, C0-C1, F5-FF}: invalid lead bytes, transition to $S_{\text{err}}$
\end{itemize}

The \texttt{80-BF*} transitions in 3-byte and 4-byte paths indicate
context-dependent continuation byte ranges that reject overlong encodings (e.g.,
\texttt{E0} requires \texttt{A0-BF} as the first continuation) and surrogate
halves (\texttt{ED} requires \texttt{80-9F}). Similarly, \texttt{F0} requires
\texttt{90-BF} and \texttt{F4} requires \texttt{80-8F} to reject codepoints
above U+10FFFF.

\section{Level 0 Trial Set Construction}
\label{app:level0-details}

The Level 0 trial set evaluates context-free UTF-8 generation by prompting the
model with isolated OOV characters under byte-fallback. We reuse the frequency-tiered
dataset described in Section~\ref{app:eval-dataset}, stratifying by
\emph{Common}, \emph{Uncommon}, \emph{Rare}, and \emph{Unseen} tiers.

For each character $c$ in the trial set, we construct the prompt by tokenizing
$c$ (which produces byte-fallback tokens) and providing the first $k$ bytes as
context. The model must generate the remaining bytes to complete a valid UTF-8
character. We sweep $k \in \{1, 2\}$ for 3-byte characters and $k \in \{1,
2, 3\}$ for 4-byte characters to evaluate partial-completion difficulty.

\section{Level 1 Prompt Construction}
\label{app:level1-details}

Level 1 evaluates context-guided byte retrieval by embedding OOV characters in
natural language sentences. Target characters are sourced from the pre-tokenized
training corpus by identifying contiguous byte-fallback sequences (e.g.,
\texttt{<0xE4><0xB8><0xAD>}) and decoding them to UTF-8.

For each target character $c$ with byte sequence $B_c = (B_p, B_r)$, where $B_p$
is the provided prefix and $B_r$ is the suffix to generate, we construct a
prompt $P = C_{\text{ctx}} \| B_p$ where $C_{\text{ctx}}$ is the preceding
sentence context. The model is evaluated on whether it emits exactly $B_r$ as
the immediate continuation.

We use a fixed split of 256 prompts per language (Japanese, Korean, Chinese) for
checkpoint-wise monitoring, with prompts stratified by character frequency to
ensure coverage across difficulty levels.

\section{Other Evaluation Metrics}

\begin{figure*}[t]
    \centering

    \begin{subfigure}[t]{0.48\textwidth}
        \centering
        \includegraphics[width=\linewidth]{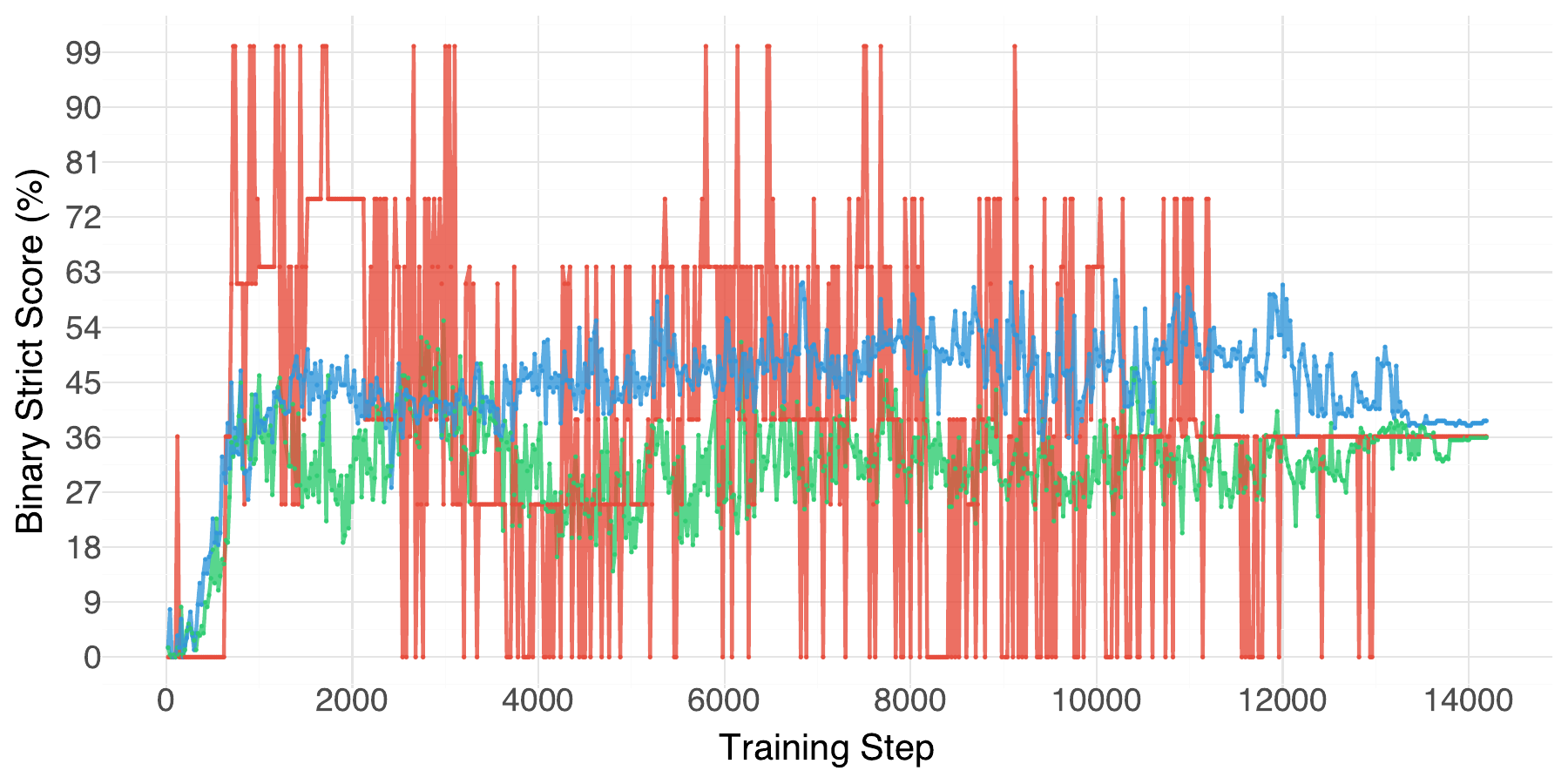}
        \caption{L0: Binary}
    \end{subfigure}
    \begin{subfigure}[t]{0.48\textwidth}
        \centering
        \includegraphics[width=\linewidth]{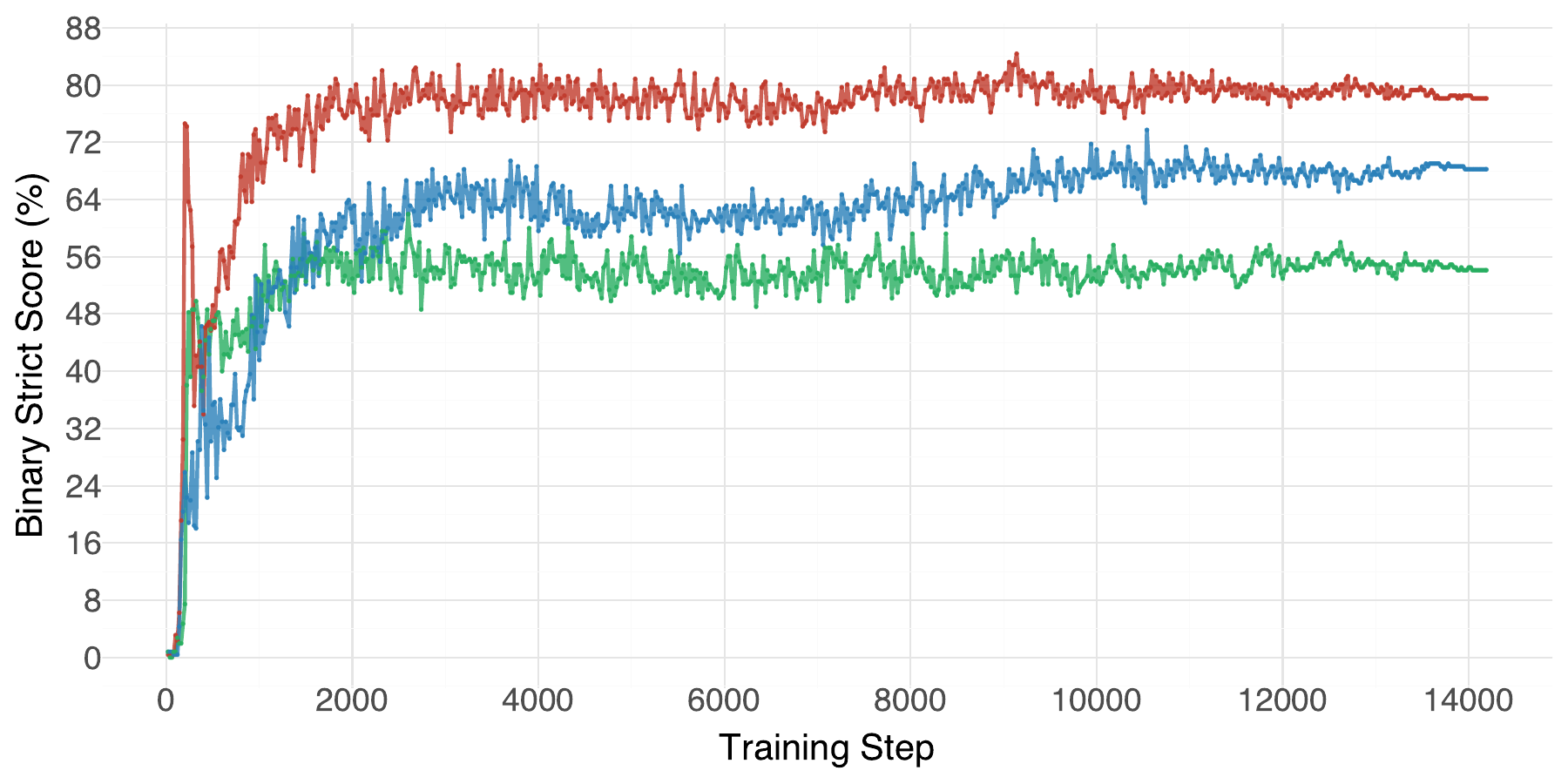}
        \caption{L1: Binary}
    \end{subfigure}

    \begin{subfigure}[t]{0.48\textwidth}
        \centering
        \includegraphics[width=\linewidth]{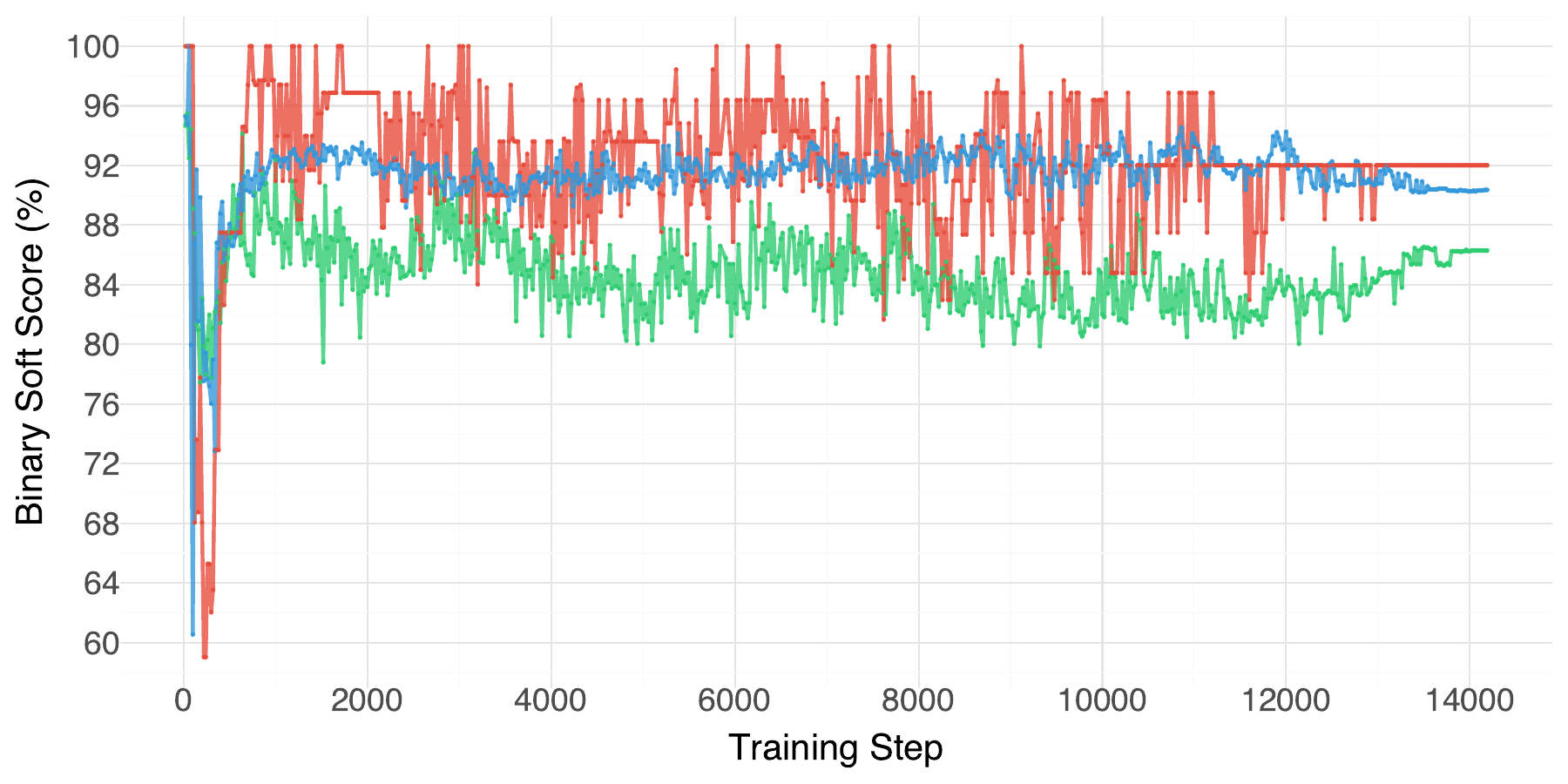}
        \caption{L0: Binary Soft}
    \end{subfigure}
    \begin{subfigure}[t]{0.48\textwidth}
        \centering
        \includegraphics[width=\linewidth]{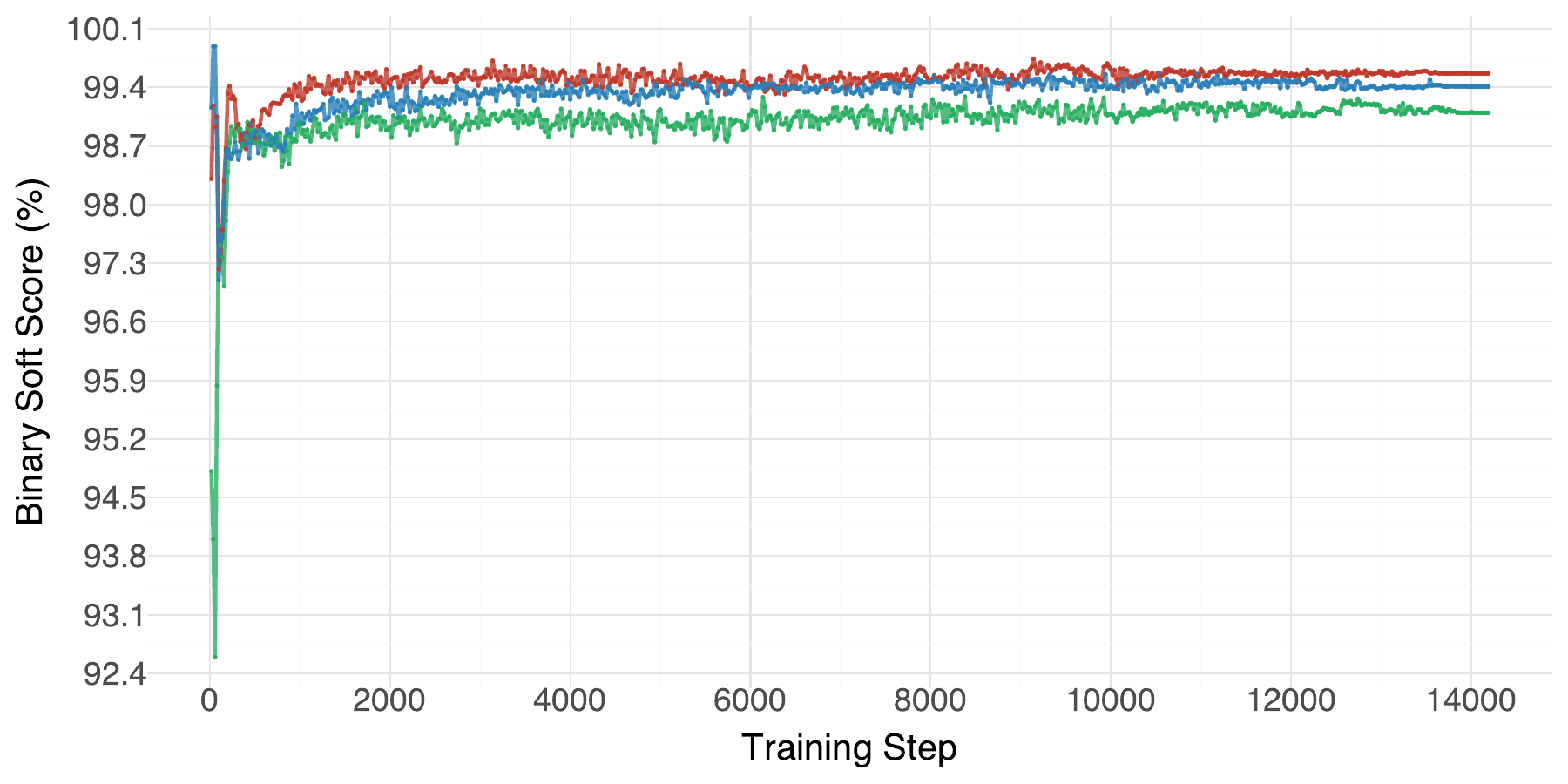}
        \caption{L1: Binary Soft}
    \end{subfigure}

    \caption{Side-by-side comparison of other validity metrics evaluated. The \textbf{left column} shows the baseline (L0) and the \textbf{right column} shows the context-guided setting (L1). Results of Chinese, Japanese, and Korean are plotted in green, red, and blue, respectively. Note how L0 in general is very unstable, while L1 is relatively more stable.}
    \label{fig:binary_grid_appendix}
\end{figure*}

At the final checkpoint (step 14,189, corresponding to 80B tokens), binary strict
validity---which requires the entire generated sequence to be valid UTF-8---was
highest on the \textit{Common} tier (50.47\%), followed by \textit{Unseen}
(33.33\%), \textit{Uncommon} (32.48\%), and \textit{Rare} (30.24\%). The binary
soft metric, which credits complete characters without penalizing trailing
incomplete bytes, showed higher scores across all tiers: \textit{Common}
(92.37\%), \textit{Uncommon} (90.18\%), \textit{Rare} (89.67\%), and
\textit{Unseen} (79.62\%).

\subsection{Binary Score}
The binary score awards credit only for completely valid sequences:
\begin{equation}
V_{\text{binary}}(B) =
\begin{cases}
1 & \text{if DFA ends in } S_0 \text{ with no errors} \\
0 & \text{otherwise}
\end{cases}
\end{equation}
This score is appropriate for final evaluation where partial progress is
insufficient.

\subsection{Binary Soft Score}
A third variant credits only complete characters without penalizing trailing
incomplete bytes:
\begin{equation}
V_{\text{binary-soft}}(B) = \frac{b_c}{|B|}
\end{equation}
This score distinguishes between invalid sequences (which reduce $b_c$) and
merely incomplete ones (which do not).

\section{Level 0 and 1: Building the Evaluation Character Set}
\label{app:eval-dataset}

To evaluate the model's zero-shot generalization on out-of-vocabulary (OOV)
characters, we constructed a stratified dataset $D_{trial}$ of 4,000 characters
representing four frequency tiers within the training corpus: \textit{Common},
\textit{Uncommon}, \textit{Rare}, and \textit{Unseen}.

We defined the universe of known characters $K$ as the union $K = V \cup S$,
where $V$ is the set of unique Unicode characters in the model's Byte-Pair
Encoding (BPE) vocabulary tokens, and $S$ is the set of unique OOV characters in
the training corpus as recorded in frequency snapshot $F$. This definition
excludes any character the model encountered during training, either as a token
or byte-fallback sequence. We obtained $|K| = 50{,}708$.

The dataset $D_{trial}$ comprises four disjoint subsets of $N=1000$ samples
each. We applied a balanced sampling strategy to ensure linguistic diversity,
targeting equal distribution of Han (CJK Ideographs), Hangul (Korean), and Kana
(Japanese) scripts, plus other symbols. The natural distribution of OOV
characters constrained this balance.

\subsection{Common Tier}

The \textbf{Common Tier ($T_{common}$)} was selected from the OOV characters
used in the Level 1 (Context-Guided) evaluation, enabling direct comparability
between Level 0 and Level 1 performance on high-frequency characters. We
prioritized the 244 characters from the Level 1 Trial set and filled the
remainder with script-balanced selection from the Level 1 source data.

\subsection{Uncommon and Rare Tiers}

For the \textbf{Uncommon Tier ($T_{uncommon}$)} and \textbf{Rare Tier
($T_{rare}$)}, we employed script-aware sampling rather than global frequency
thresholds to ensure linguistic diversity. The global distribution of OOV
character frequencies is skewed, with CJK Ideographs dominating the long tail.
To prevent mono-script tiers, we stratified selection by script type. For the
Rare Tier, we selected the 300 lowest-frequency characters for each of Han and
Hangul, and 100 for Other. For Kana, we included all 28 OOV characters from the
corpus snapshot, as most Kana are vocabulary-covered. For the Uncommon Tier, we
selected the next-lowest 300 characters for Han and Hangul from the remaining
pool. This ensures evaluation of the rarest Hangul even though Hangul characters
generally appear more frequently than rare Han characters.

\subsection{Unseen Tier}

The \textbf{Unseen Tier ($T_{unseen}$)} tests zero-shot generalization. We
defined candidate universe $U$ comprising all Unicode codepoints in the CJK
Unified Ideographs (including Extensions A and B) and Hangul Syllables blocks.
The pool of candidate unseen characters was $P_{unseen} = U \setminus K$,
yielding 42,871 candidates never observed in training. Due to high coverage of
Basic Multilingual Plane (BMP) characters in training, this pool consists
predominantly of CJK Extension B characters (codepoints above U+10000), which
require 4-byte UTF-8 encoding rather than the 3-byte encoding used by BMP
characters in the other tiers. This means the Unseen tier tests generalization
to a different UTF-8 byte-pattern family (\texttt{11110xxx} start bytes) in
addition to novel character identity. The final sample contains 988 Han
characters and 12 Hangul characters.

\subsection{Dataset Interleaving}

The final dataset $D_{trial}$ was constructed using deterministic interleaving
across the four tiers. The sequence follows the repeating pattern $[T_{common},
T_{uncommon}, T_{rare}, T_{unseen}, \dots]$. The interleaved structure
guarantees balanced coverage across frequency tiers for each language.

\paragraph{Note on computational limitations.}
Due to computational resource constraints, the experiments reported in the main
text use a subset of $M=256$ samples per language group (64 per frequency tier).
A full sweep over the complete 4,000-character dataset is left for future work.

\section{Level 1: Synthetic Data Generation}
\label{app:gemini-contexts}

To avoid data leakage from re-using pre-training text, we generate synthetic
sentence contexts using Gemini~3~Pro. For each target OOV character $c$ and
language $L$, we prompt the model:

\begin{quote}
\textit{Write a single grammatically correct sentence in [L] that naturally
incorporates the character ``[c]''. The sentence should be 10--20 words and
provide clear semantic context for the character.}
\end{quote}

Generated sentences are filtered for: (1) presence of the target character, (2)
correct language identification via langdetect, (3) uniqueness (no duplicate
sentences across the dataset), and (4) length constraints (10--30 tokens after
BPE tokenization). Approximately 15\% of generated sentences are rejected by
these filters.

\section{Cross-Model Evaluation Details}
\label{app:cross-model-details}

The cross-model results in Section~\ref{sec:eval-framework} use publicly released
instruction-tuned checkpoints retrieved from the Hugging Face Hub. We use the
default tokenizer shipped with each checkpoint and rely on the model's native
byte handling: SentencePiece-based models (Llama-2, Llama-3.2, Mistral, our
baseline) expose explicit byte-fallback tokens, whereas GPT-2-style BPE models
(Gemma-3, OLMo-2, Qwen-3.5) represent arbitrary bytes through their pre-tokenizer
without dedicated byte-fallback tokens. For every model we feed the \emph{same}
UTF-8 byte prefixes derived from the Level~0 and Level~1 trial sets; the model
is then asked to continue from the corresponding tokenization of that prefix.
We use greedy decoding throughout, generate up to 5 tokens per trial, and
average all reported values across languages and prefix lengths. All 100 trial
samples per language are reused unchanged from the baseline protocol
(Appendices~\ref{app:level0-details} and~\ref{app:level1-details}). Decoded byte
streams are scored by the same UTF-8 DFA used for the baseline; partial-credit
validity, strict validity, and Term Match are computed identically.

\section{Distractor Token Distribution at \boldmath$\Delta_{LL}>0$}
\label{app:distractor}

At the final checkpoint, Level~1 has 51 cases where $\Delta_{LL}>0$ (the model
assigns higher teacher-forced likelihood to the gold continuation) but greedy
decoding emits a different byte. All 51 emitted tokens are byte tokens in the
range \texttt{0x80}--\texttt{0xBF}; no subword token and no lead byte appears.
Table~\ref{tab:distractor-leads} groups the emissions by lead byte, and
Table~\ref{tab:distractor-bytes} reports the full distribution of emitted
bytes.

\begin{table}[h]
\centering
\small
\begin{minipage}[t]{0.58\linewidth}
\centering
\caption{For three lead bytes the model emits a single continuation byte
across all $\Delta_{LL}>0$ failures. Mode collapse is exact within each lead
byte.}
\label{tab:distractor-leads}
\begin{tabular}{@{}lll@{}}
\toprule
Lead byte & Emitted (count) & Produces \\
\midrule
\texttt{0xE3} (JA)   & \texttt{0x80} (8/8) & CJK Symbols U+3000 \\
\texttt{0xEC} (KO)   & \texttt{0x97} (4/4) & Common Hangul \\
\texttt{0xF0} (4B)   & \texttt{0x9F} (3/3) & Emoji range \\
\bottomrule
\end{tabular}
\end{minipage}\hfill
\begin{minipage}[t]{0.38\linewidth}
\centering
\caption{Full distribution of emitted bytes for the 51 $\Delta_{LL}>0$
failures at the final checkpoint. All values lie in \texttt{0x80}--\texttt{0xBF}
(UTF-8 continuation byte range), so the structural-validity DFA accepts every
emission. The semantic failure is a mode collapse in $P(\text{byte}_2 \mid
\text{byte}_1)$.}
\label{tab:distractor-bytes}
\begin{tabular}{@{}lr@{}}
\toprule
Emitted byte & Count \\
\midrule
\texttt{0x80} & 12 \\
\texttt{0xB7} & 5 \\
\texttt{0x97} & 4 \\
\texttt{0x8B} & 4 \\
\texttt{0x8C} & 3 \\
\texttt{0x9F} & 3 \\
\texttt{0xB9} & 2 \\
Other (18 singletons) & 18 \\
\midrule
Total & 51 \\
\bottomrule
\end{tabular}
\end{minipage}
\end{table}

\end{document}